\documentclass[letterpaper]{article} % DO NOT CHANGE THIS
\usepackage{aaai2026}  % DO NOT CHANGE THIS
\usepackage{times}  % DO NOT CHANGE THIS
\usepackage{helvet}  % DO NOT CHANGE THIS
\usepackage{courier}  % DO NOT CHANGE THIS
\usepackage[hyphens]{url}  % DO NOT CHANGE THIS
\usepackage{graphicx} % DO NOT CHANGE THIS
\usepackage{natbib}  % DO NOT CHANGE THIS AND DO NOT ADD ANY OPTIONS TO IT
\usepackage{caption} % DO NOT CHANGE THIS AND DO NOT ADD ANY OPTIONS TO IT
\usepackage{algorithm}
\usepackage{algorithmic}
\usepackage{pifont}
\usepackage{xcolor}
\usepackage{booktabs}
\newcommand{\cmark}{\textcolor{black}{\ding{51}}} % √ green!60!black
 
\usepackage{makecell}
\usepackage{amsmath}
\usepackage{amssymb}
\usepackage{multirow}

\usepackage{newfloat}
\usepackage{listings}
\DeclareCaptionStyle{ruled}{labelfont=normalfont,labelsep=colon,strut=off} % DO NOT CHANGE THIS
\floatstyle{ruled}
\newfloat{listing}{tb}{lst}{}
\floatname{listing}{Listing}
\title{UniFit: Towards Universal Virtual Try-on with \\ MLLM-Guided Semantic Alignment}
\author{
    Wei Zhang\textsuperscript{\rm 1,}\equalcontrib,
    Yeying Jin\textsuperscript{\rm 2,}\equalcontrib,
    Xin Li\textsuperscript{\rm 3},
    Yan Zhang\textsuperscript{\rm 4},\\
    Xiaofeng Cong\textsuperscript{\rm 5},
    Cong Wang\textsuperscript{\rm 6},
    Fengcai Qiao\textsuperscript{\rm 7},
    Zhichao Lian\textsuperscript{\rm 1,}\thanks{Corresponding Author}
}
\affiliations{
    \textsuperscript{\rm 1}School of Cyber Science and Engineering, Nanjing University of Science and Technology\\
    \textsuperscript{\rm 2}National University of Singapore\\
    \textsuperscript{\rm 3}University of Science and Technology of China\\
    \textsuperscript{\rm 4}	ByteDance\\
    \textsuperscript{\rm 5}	Southeast University\\
    \textsuperscript{\rm 6}	University of California, San Francisco\\
    \textsuperscript{\rm 7}	National University of Defense Technology\\
    
    zwplus\_pro@njust.edu.cn, jinyeying@u.nus.edu, xin.li@ustc.edu.cn, yanzhang.yz@bytedance.com \\
    cxf\_svip@163.com, supercong94@gmail.com, fcqiao@nudt.edu.cn,
    newlzcts@njust.edu.cn
}
\usepackage{bibentry}

\begin{document}

\maketitle

\begin{abstract}
Image-based virtual try-on (VTON) aims to synthesize photorealistic images of a person wearing specified garments. Despite significant progress, building a universal VTON framework that can flexibly handle diverse and complex tasks remains a major challenge. Recent methods explore multi-task VTON frameworks guided by textual instructions, yet they still face two key limitations: (1) semantic gap between text instructions and reference images, and (2) data scarcity in complex scenarios. To address these challenges, we propose UniFit, a universal VTON framework driven by a Multimodal Large Language Model (MLLM). Specifically, we introduce an MLLM-Guided Semantic Alignment Module (MGSA), which integrates multimodal inputs using an MLLM and a set of learnable queries. By imposing a semantic alignment loss, MGSA captures cross-modal semantic relationships and provides coherent and explicit semantic guidance for the generative process, thereby reducing the semantic gap. Moreover, by devising a two-stage progressive training strategy with a self-synthesis pipeline, UniFit is able to learn complex tasks from limited data. Extensive experiments show that UniFit not only supports a wide range of VTON tasks, including multi-garment and model-to-model try-on, but also achieves state-of-the-art performance. The source code and pretrained models are available at \url{https://github.com/zwplus/UniFit}.

\end{abstract}


\section{Introduction}

Image-based virtual try-on (VTON) aims to synthesize photorealistic images of a person wearing specified garments, with broad applications in e-commerce and digital content creation. Recent methods, such as OOTD~\cite{xu2025ootdiffusion}, IMAGDressing~\cite{shen2025imagdressing}, and TryoffDiff~\cite{velioglu2024tryoffdiff}, have achieved impressive results. However, most existing approaches are task-specific, such as single-garment try-on. 

\begin{figure}[t]
\centering
\includegraphics[width=0.99\columnwidth]{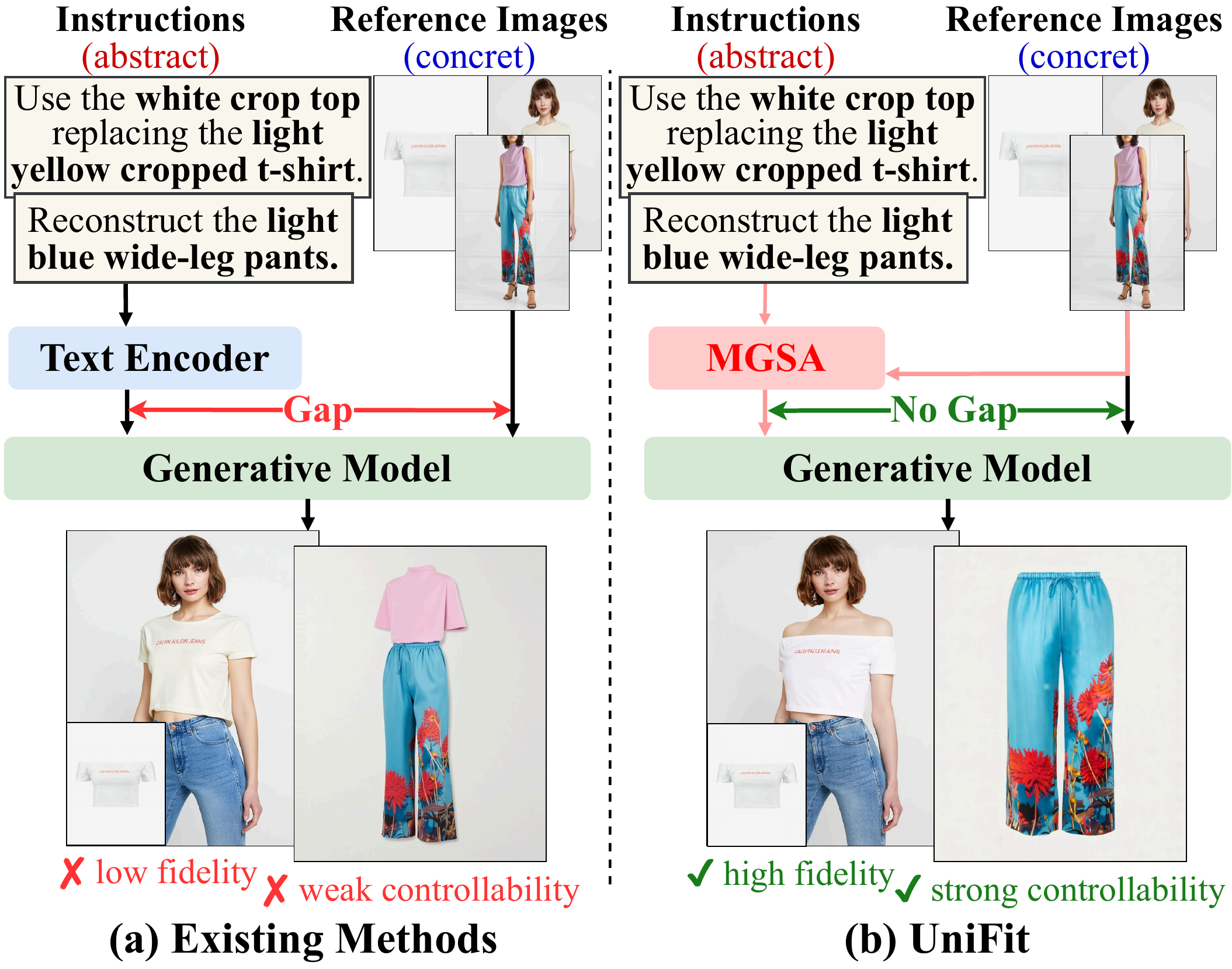}
\caption{Motivation of UniFit: (a) Existing instruction-guided VTON methods process text and images separately, resulting in a semantic gap. (b) Our UniFit introduces an MLLM-Guided Semantic Alignment Module (MGSA), which integrates textual and visual inputs to produce coherent and explicit semantic guidance for the generative model, effectively bridging the semantic gap.}
\label{fig1}
\end{figure}

\begin{table*}[t]
\begin{center}
\begin{small}
\begin{tabular}{lcccccc}
\toprule
\textbf{Method} & \makecell{Single-garment \\ try-on} & \makecell{Model-free \\ try-on} & \makecell{Garment \\ reconstruction} & \makecell{Multi-view \\ try-on} & \makecell{Multi-garment \\ try-on} & \makecell{Model-to-model \\ try-on} \\
\midrule
AnyFit~\cite{li2024anyfit}       & \cmark & - & - & - & \cmark & - \\
CatVTON~\cite{chong2024catvton}      & \cmark & - & - & - & - & \cmark \\
MV-VTON~\cite{wang2025mv}      & \cmark & - & - & \cmark & - & - \\
Any2AnyTryon~\cite{guo2025any2anytryon} & \cmark & \cmark & \cmark & - & - & - \\
\midrule
UniFit (ours) & \cmark & \cmark & \cmark & \cmark & \cmark & \cmark \\
\bottomrule
\end{tabular}
\end{small}
\end{center}
\caption{Comparison of VTON functionalities achieved by UniFit and previous methods. UniFit is capable of handling multiple VTON tasks, including multi-view try-on, multi-garment try-on, and model-to-model try-on, surpassing previous methods.}
\label{task_table}
\end{table*}

To address this limitation, recent studies~\cite{guo2025any2anytryon,zhang2024mmtryon} have begun to explore multi-task VTON frameworks guided by textual instructions. 
As illustrated in Figure~\ref{fig1}(a), a text encoder (e.g., CLIP~\cite{radford2021learning} or T5~\cite{raffel2020exploring}) extracts task-relevant cues which condition the generative model to extract relevant visual features from the reference images. However, relying only on abstract language for guidance often results in poor grounding in concrete visual details (e.g., texture or logo shape), causing a semantic gap. This gap prevents the model from faithfully integrating task-relevant visual features from the reference images, often resulting in outputs with low fidelity and weak controllability. Moreover, limited data for complex scenarios in public datasets further restricts current methods from supporting advanced tasks such as model-to-model and multi-garment try-on (see Table~\ref{task_table}).

To address these challenges, we propose UniFit, an instruction-guided, universal VTON framework that deeply integrates a Multimodal Large Language Model (MLLM) with a Diffusion Transformer (DiT)~\cite{peebles2023scalable}. First, to align textual instructions with reference images, we introduce a novel MLLM-Guided Semantic Alignment Module (MGSA). This module leverages the MLLM and a set of learnable queries to fuse multimodal inputs. Supervised by a semantic alignment loss, MGSA captures cross-modal semantic relationships and generates coherent, explicit guidance for the DiT, effectively bridging the gap between text and vision. To further enhance generation quality, we introduce a spatial attention focusing loss for the DiT module, which regularizes the cross-attention maps between the reference images and the generated output. This encourages the DiT to focus on the most task-relevant regions (e.g., the garment area to be replaced), enabling faithful transfer of fine-grained visual details.

To address data scarcity towards universal VTON, we propose a two-stage progressive training strategy with a self-synthesis pipeline as illustrated
in Figure~\ref{fig3}. In the first stage, a Base Model is trained on publicly available datasets to learn foundational tasks such as model-free try-on and garment reconstruction. Then, the Base Model serves as a data synthesizer, generating high-quality paired samples for advanced tasks. For instance, it leverages its learned garment reconstruction ability to extract disentangled upper and lower garments from a single image, creating training samples for multi-garment try-on. In the second stage, the model is fine-tuned on a composite dataset of both real and synthesized data to support more advanced tasks. This training strategy enables UniFit to gradually master a wide range of VTON tasks, effectively overcoming the data limitations of existing benchmarks. Our contributions are summarized as follows:
\begin{itemize}
    \item We propose UniFit, a universal VTON framework capable of handling a diverse range of tasks, from standard single-garment try-on to complex multi-garment and model-to-model try-on.

    \item We propose an MLLM-Guided Semantic Alignment Module (MGSA) to bridge the semantic gap between textual instructions and reference images. By imposing a semantic alignment loss, MGSA can capture cross-modal relationships and generate explicit guidance for the generative process. Additionally, we introduce a spatial attention focusing loss to regularize the DiT’s attention, ensuring faithful transfer of fine-grained visual details.
    
    \item We devise a two-stage progressive training strategy with a self-synthesis pipeline, which effectively overcomes data scarcity for advanced VTON tasks, significantly enhancing the model's ability to handle multi-garment and model-to-model try-on tasks.
\end{itemize}

\begin{figure*}[t]
\centering
\includegraphics[width=0.99\textwidth]{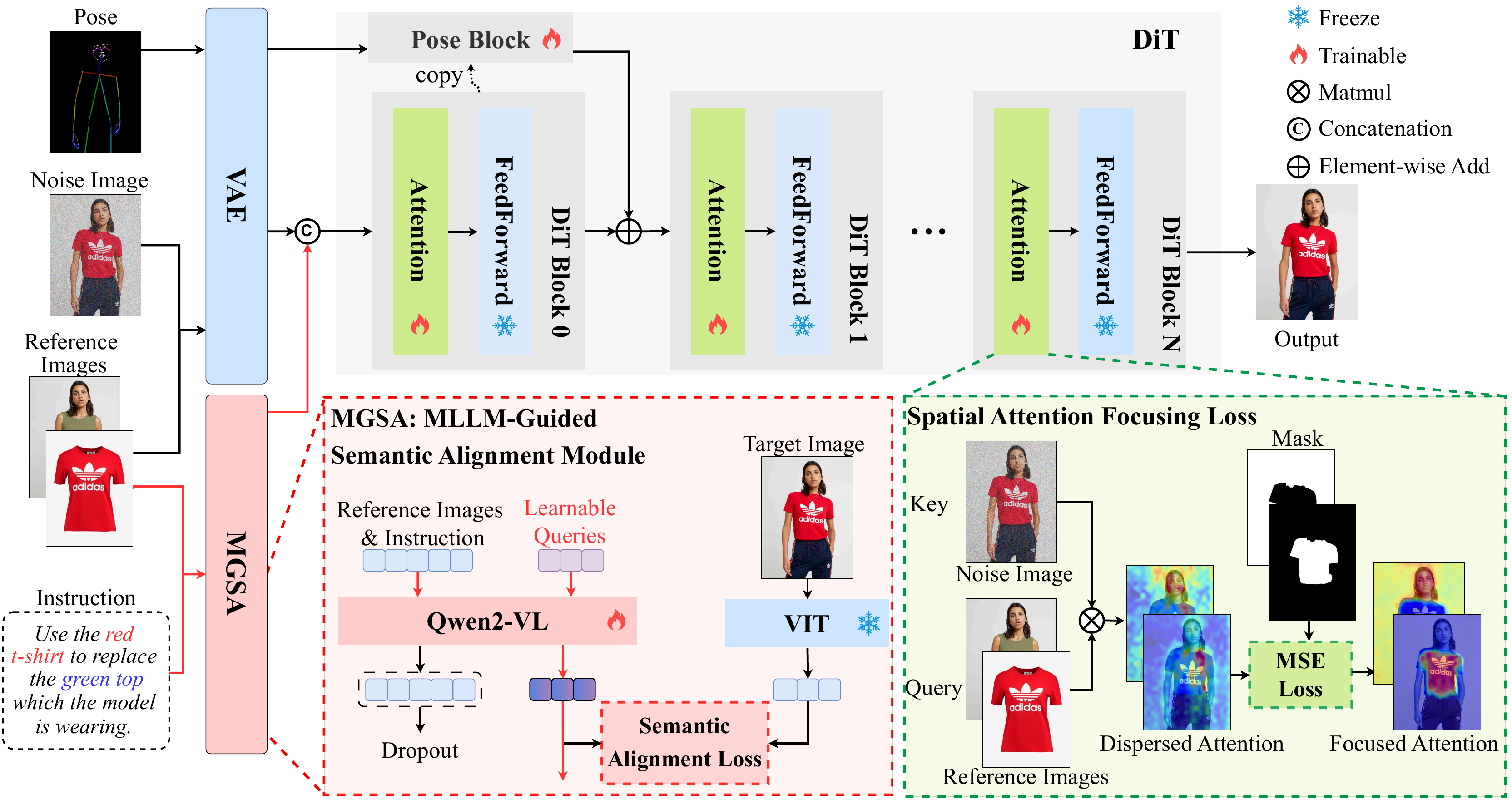}
\caption{\textbf{Overview of UniFit.} UniFit consists of three main components: the MGSA module (red), the DiT (gray), and the VAE encoder (blue). The MGSA encodes multimodal inputs into coherent semantic guidance. The VAE extracts low-level visual features from reference images. The DiT generates the output image conditioned on the semantic guidance and low-level visual features. Additionally, a spatial attention focusing loss (green) supervises the attention maps of the DiT, encouraging the model to focus on the most task-relevant regions (e.g., the try-on area in single-garment try-on task).
}
\label{fig2}
\end{figure*}

\section{Method}
\subsubsection{Overview }
We propose UniFit, a universal virtual try-on framework capable of generating corresponding try-on results based on diverse textual instructions and reference images. As illustrated in Figure~\ref{fig2}, the overall architecture consists of three main components: an MLLM-Guided Semantic Alignment Module (MGSA) to associate textual instructions with visual content, a VAE encoder~\cite{kingma2013auto} for extracting low-level visual features, and a Diffusion Transformer (DiT) serving as the core generative backbone. The generation pipeline begins with two parallel streams. The MGSA leverages a pre-trained MLLM (Qwen2-VL\cite{wang2024qwen2}) and a set of learnable queries to capture the semantic relationships between textual instructions and reference images, producing a high-level semantic representation of the target image, denoted as $T_q$. This representation serves as explicit semantic guidance for the subsequent generation process. Simultaneously, the VAE encoder processes the reference images to extract fine-grained visual features, denoted as $r = \{r_1, \ldots, r_n\}$, which provide rich visual cues for generation. We then concatenate $T_q$, the noisy latent $z_t$ of the target image, and the reference tokens $r$ to form the DiT input $[T_q; z_t; r_1; \ldots; r_n]$. During the iterative denoising process, the DiT, guided by $T_q$, effectively integrates the fine-grained visual information in $r$ to progressively refine $z_t$ into the final output image.  

To enhance pose consistency and detail fidelity, we incorporate two auxiliary components. First, inspired by Catv2ton~\cite{chong2025catv2ton}, we employ a Pose Block (a trainable copy of the first DiT block) to extract pose features, which are then injected into the second DiT block's input via element-wise addition to enforce pose alignment. Second, a spatial attention focusing loss regularizes the DiT's attention maps, ensuring the model precisely focuses on relevant regions when transferring details. During training, we jointly optimize a specific set of components: the MGSA (including the MLLM and learnable queries), the attention layers within the DiT, and the Pose Block, while keeping other parts like the VAE and the DiT's FeedForward layers frozen. The overall training objective is a combination of three components: (1) a flow matching loss, (2) the semantic alignment loss, and (3) the spatial attention focusing loss.

\begin{figure*}[t]
\centering
\includegraphics[width=0.99\textwidth]{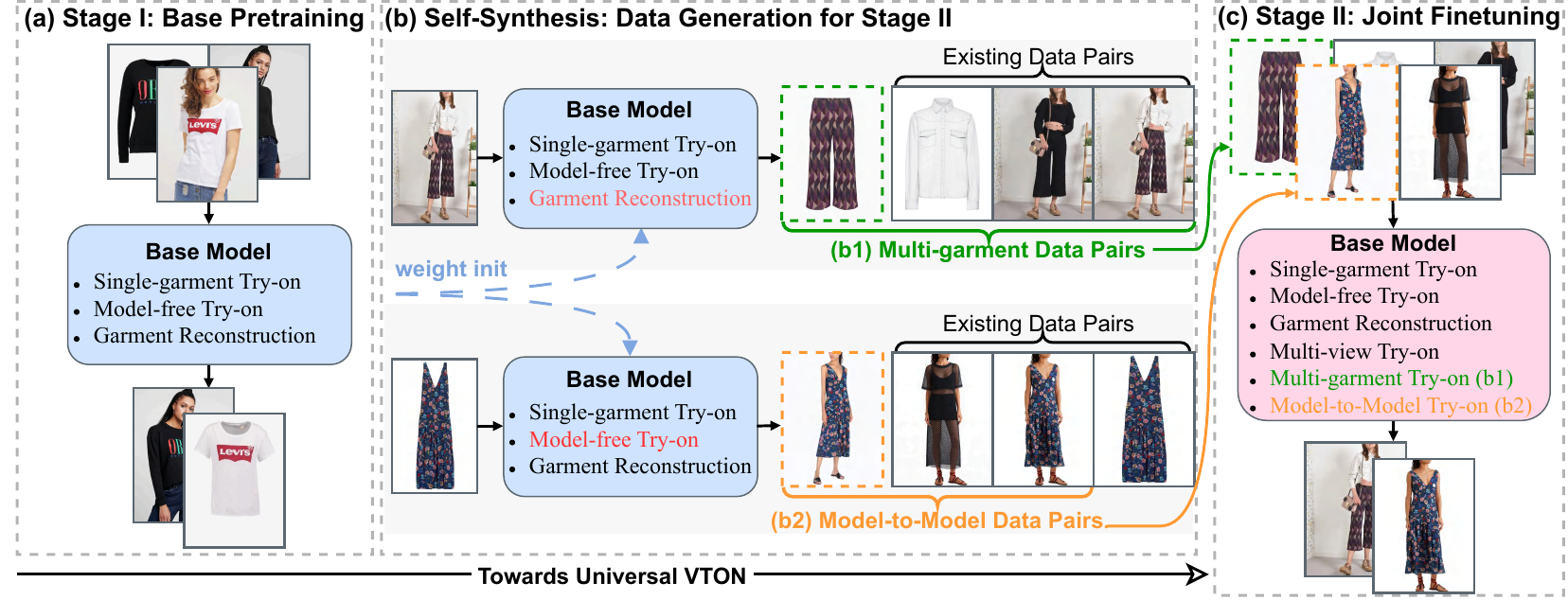}
\caption{\textbf{Towards Universal VTON}: Two-Stage Progressive Training Strategy of UniFit with Self-Synthesis.
\textbf{(a) Stage I:} A Base Model is trained on foundational tasks using public datasets. 
\textbf{(b) Self-Synthesis:} The trained model is then used to generate pseudo-paired data for complex scenarios. 
\textbf{(b1)} For multi-garment try-on, we reconstruct garments from full-body images. 
\textbf{(b2)} For model-to-model try-on, we synthesize new person images conditioned on given garments. 
\textbf{(c) Stage II:} The model is fine-tuned on a composite dataset of both real and synthesized samples, enabling generalization to a wide range of VTON tasks.}
\label{fig3}
\end{figure*}

\subsection{MLLM-Guided Semantic Alignment Module}

A key challenge in instruction-guided multi-task VTON frameworks is bridging the semantic gap between abstract textual instructions and concrete visual content (see Figure~\ref{fig1}). To address this issue, we propose the MLLM-Guided Semantic Alignment Module (MGSA) (see Figure~\ref{fig2} red), which leverages a pre-trained Multimodal Large Language Model (Qwen2-VL) to jointly process both textual and visual inputs. To enable the MGSA to effectively capture the semantic relationships between these inputs and generate a coherent and explicit semantic guidance for the subsequent generation process, we introduce two core components: a set of learnable queries and a semantic alignment loss.

As illustrated in Figure~\ref{fig2}, the MGSA jointly processes reference image tokens and textual instruction tokens via the pre-trained Qwen2-VL. However, the resulting token sequence is typically long, introducing significant redundant information and computational overhead for the downstream DiT. To mitigate this, we introduce a set of learnable queries, $T_q \in \mathbb{R}^{N_q \times D_q}$, where $N_q$ is the number of queries and $D_q$ denotes their dimensionality. These queries are appended to the end of Qwen2-VL’s input sequence and serve as information aggregators. Through Qwen2-VL's causal attention mechanism, they distill task-relevant signals from the extensive multimodal sequence into a compact representation.

To ensure that the representation $T_q$ is semantically meaningful and aligned with the target output, we introduce a semantic alignment loss, $\mathcal{L}_{\text{align}}$, which explicitly aligns $T_q$ with the ground-truth visual representation of the target image. Specifically, we extract a set of visual tokens $ T_v \in \mathbb{R}^{N_v \times D_v}$ from the ground-truth target image using a frozen VIT~\cite{dosovitskiy2020image}, and set $N_q = N_v$. We then enforce a token-wise alignment between them using cosine similarity:
\begin{equation}
\mathcal{L}_{\text{align}} = - \frac{1}{N_v} \sum_{n=1}^{N_v} \cos\left(T_{v,n}, \text{MLP}(T_{q,n})\right),
\end{equation}
where $T_{v,n}$ and $T_{q,n}$ represent the $n$-th tokens from the visual and query embeddings, respectively. A lightweight MLP projection head is used to match the feature dimensions of $T_q$ and $T_v$. Together, the learnable queries and semantic alignment loss enable the MGSA to fuse multimodal inputs into a coherent semantic representation that captures task intent and visual cues. This representation provides explicit guidance to the DiT, allowing it to aggregate relevant features and generate high-quality outputs aligned with the instructions.

\subsection{Spatial Attention Focusing Loss}
Although the MGSA provides strong high‑level semantic guidance, the DiT’s attention still tends to be dispersed across irrelevant regions in both the reference image and the generated output. This dispersion often degrades fine details and introduces visual artifacts. As illustrated in the bottom‑right corner of Figure~\ref{fig2} (see green), the initial cross‑attention between the garment image and the output is scattered rather than concentrated on the intended try‑on area. To mitigate this issue, we introduce a spatial attention focusing loss inspired by DreamO~\cite{mou2025dreamo}. The loss explicitly regularizes the cross‑attention maps, encouraging the model to focus on regions that are truly critical for the task.

In detail, we first compute the cross‑attention map $AttnMap \in \mathbb{R}^{l_{r_i} \times l_{z_t}}$, using the reference tokens as queries and the output tokens as keys, where $l_{r_i}$ and $l_{z_t}$ denote their respective sequence lengths. Then, for try-on tasks, we average the map over the reference‑token axis to obtain an output‑centric response map $M \in \mathbb{R}^{l_{z_t}}$, which highlights the target regions expected to receive garment features or the model’s appearance features. For garment‑reconstruction tasks, we average over the output‑token axis, producing a reference‑centric map $M \in \mathbb{R}^{l_{r_i}}$ that pinpoints where to extract critical garment details in the reference image. Specifically, the model-to-model try-on task requires both extracting a specific garment from the reference image and transferring it to the try-on region of the generated output. Therefore, we compute and supervise both attention response maps simultaneously. Finally, we align each task‑specific map $M$ with a ground‑truth spatial mask $M_{\text{target}}$ via an MSE loss:
\begin{equation}
\mathcal{L}_{\text{focus}} = \frac{1}{N_R \times N_L} \sum_{j=1}^{N_L} \sum_{i=1}^{N_R} \left\| M_i^j - M_{\text{target},i} \right\|_2^2,
\end{equation}
where $M_i^j$ denotes the response map for the $i$-th reference image in the $j$-th attention layer, $M_{\text{target},i}$ is the ground-truth spatial mask corresponding to the $i$-th reference image, and $N_R$, $N_L$ represent the number of reference images and attention layers, respectively. This targeted supervision enforces accurate spatial correspondence, enabling high‑fidelity transfer of fine‑grained details and preventing visual artifacts in the synthesized images.

\subsection{Progressive Training via Self-Synthesis}
Public VTON datasets such as VITON-HD~\cite{choi2021viton} and DressCode~\cite{morelli2022dress} have greatly advanced the open‑source research community, yet they also exhibit inherent limitations. They provide only image pairs of a single garment and its corresponding try‑on result, which introduces two challenges. First, such pairings compel many methods to depend on garment masks. Second, and more critically, they lack sufficient data to support complex tasks such as multi-garment and model-to-model try-on. Recent mask‑free approaches mitigate the first issue by synthesizing triplet training samples via image inpainting, but data scarcity for complex tasks remains a bottleneck. To address this, we propose a two‑stage progressive training strategy with a self‑synthesis pipeline (Figure~\ref{fig3}). Throughout training, reference‑model images are generated by inpainting, thereby removing any reliance on garment masks.

\textbf{Stage I: Base Pretraining.}
In the first stage (Figure~\ref{fig3}(a)), we focus on pretraining a Base Model on three fundamental tasks: single-garment try-on, garment reconstruction, and model-free try-on. This pretraining, conducted on standard datasets such as VITON-HD and DressCode, equips the model with strong consistency-preserving capabilities and robust visual priors.

\textbf{Self-Synthesis for Complex Tasks.} Next, we use the Base Model as a data synthesizer to generate training samples for complex VTON tasks (Figure~\ref{fig3}(b)). Specifically:
\begin{itemize}
    \item \textbf{For Multi-garment Try-on (Figure~\ref{fig3}(b1)):} We leverage the model's garment reconstruction ability to extract high-quality tops or bottoms from full-body images, thus creating new training samples for this task.
    
    \item \textbf{For Model-to-model Try-on (Figure~\ref{fig3}(b2)):} We utilize the model-free try-on capability to synthesize new person images conditioned on existing garments, thereby expanding data support for the model-to-model task.
\end{itemize}
To guarantee the fidelity of the synthesized data, all synthesized samples undergo a two‑step filter: DreamSim~\cite{fu2023dreamsim} for perceptual similarity, followed by a consistency check with Qwen2.5‑VL‑7B-Instruct~\cite{bai2025qwen2}.

\begin{table}[t]
\begin{small}
\setlength{\tabcolsep}{6.5pt}
\centering
\begin{tabular}{lcccc}
\toprule
\textbf{Method} & \textbf{SSIM~$\uparrow$} & \textbf{LPIPS~$\downarrow$} & \textbf{DISTS~$\downarrow$} & \textbf{FID~$\downarrow$} \\
\midrule
TryOffDiff       & \textbf{0.792} & 0.337 & 0.227 & 21.40  \\
Any2AnyTryon & 0.762 & 0.367 & 0.231 & 13.57 \\
Ours & 0.775 & \textbf{0.281} & \textbf{0.202} & \textbf{12.58}  \\
\bottomrule
\end{tabular}
\caption{Quantitative comparison of garment reconstruction on the VITON-HD dataset. Best results are in \textbf{bold}.}
\label{Garment Reconstruction}
\end{small}
\end{table}

\begin{figure}[t]
\centering
\includegraphics[width=0.99\columnwidth]{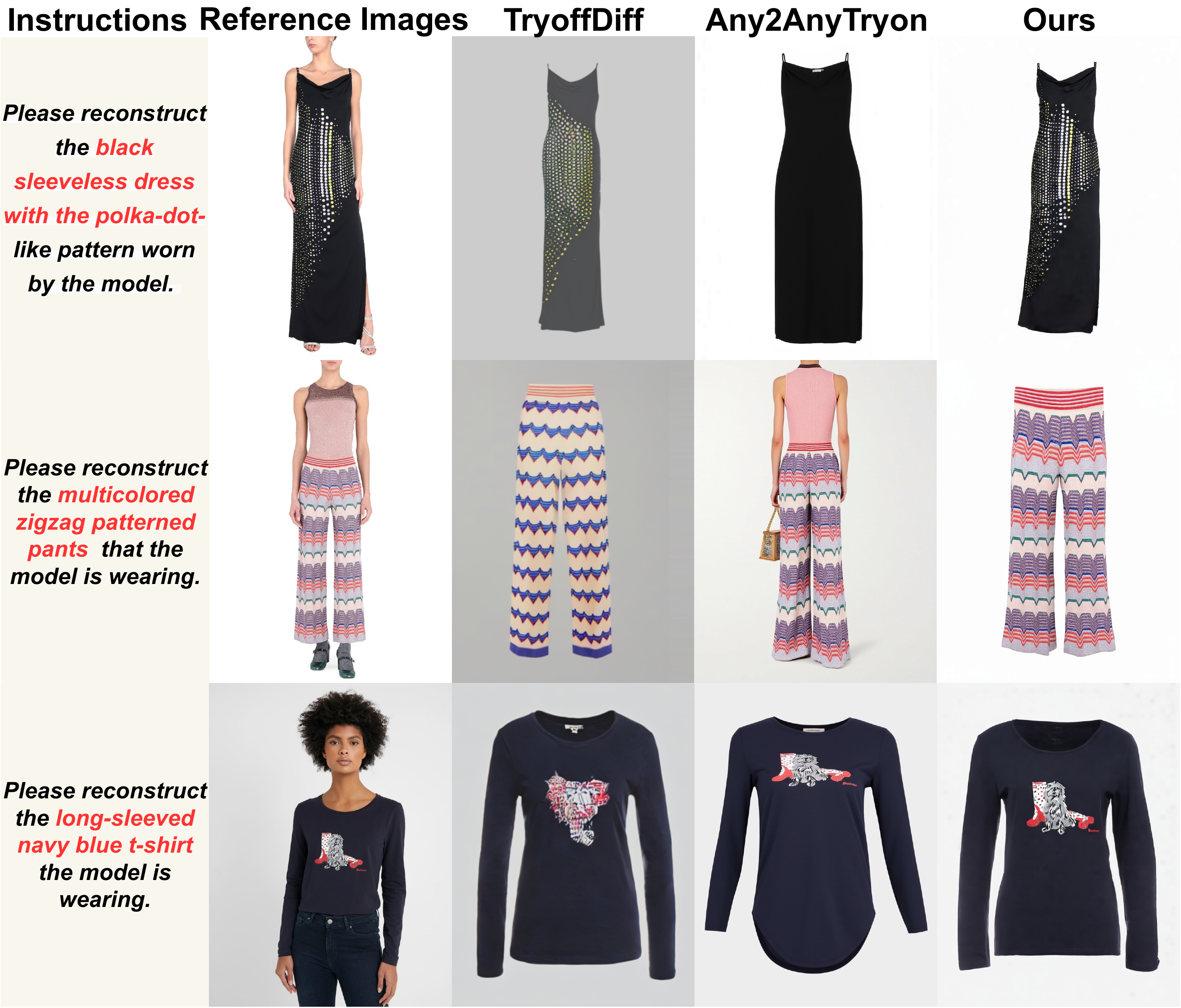}
\caption{Qualitative comparison of garment reconstruction.}
\label{fig Garment Reconstruction}
\end{figure}

\subsubsection{Stage II: Joint Finetuning.}
In the second stage (Figure~\ref{fig3}(c)), we finetune the Base Model on a composite dataset that merges real images with our high‑quality synthetic data. This joint training enables UniFit to generalize across six VTON tasks, by adding supervision for multi-garment, model-to-model, and multi-view try-on, alongside the three foundational tasks. For the multi‑view task, we construct training pairs from the MVG~\cite{wang2025mv} dataset and a manually annotated subset of 2,000 high‑quality image pairs from IG‑Pairs~\cite{shen2025imagdressing}, each containing multi‑view try‑on results and the corresponding front–back garment references. This training strategy not only mitigates the data scarcity problem but also simplifies the learning process by leveraging the visual priors acquired during Base Model pretraining.

\begin{table}[t]
\begin{small}
\setlength{\tabcolsep}{7.5pt}
\centering
\begin{tabular}{lcccc}
\toprule
\textbf{Method} & \textbf{SSIM} $\uparrow$ & \textbf{LPIPS} $\downarrow$ & \textbf{FID} $\downarrow$ & \textbf{KID} $\downarrow$ \\
\midrule
CatVTON & 0.888 & 0.075 & 9.128 & 1.130 \\
FitDiT & \textbf{0.895} & 0.067 & 9.326 & 0.913 \\
Any2AnyTryon & 0.839 & 0.088 & 8.965 & 0.981 \\
Ours & 0.883 & \textbf{0.065} & \textbf{8.799} & \textbf{0.702} \\
\bottomrule
\end{tabular}

\caption{Quantitative comparison of single garment try-on on the VITON-HD dataset. }
\label{tryon}
\end{small}
\end{table}
\begin{figure}[t]
\centering
\includegraphics[width=0.99\columnwidth]{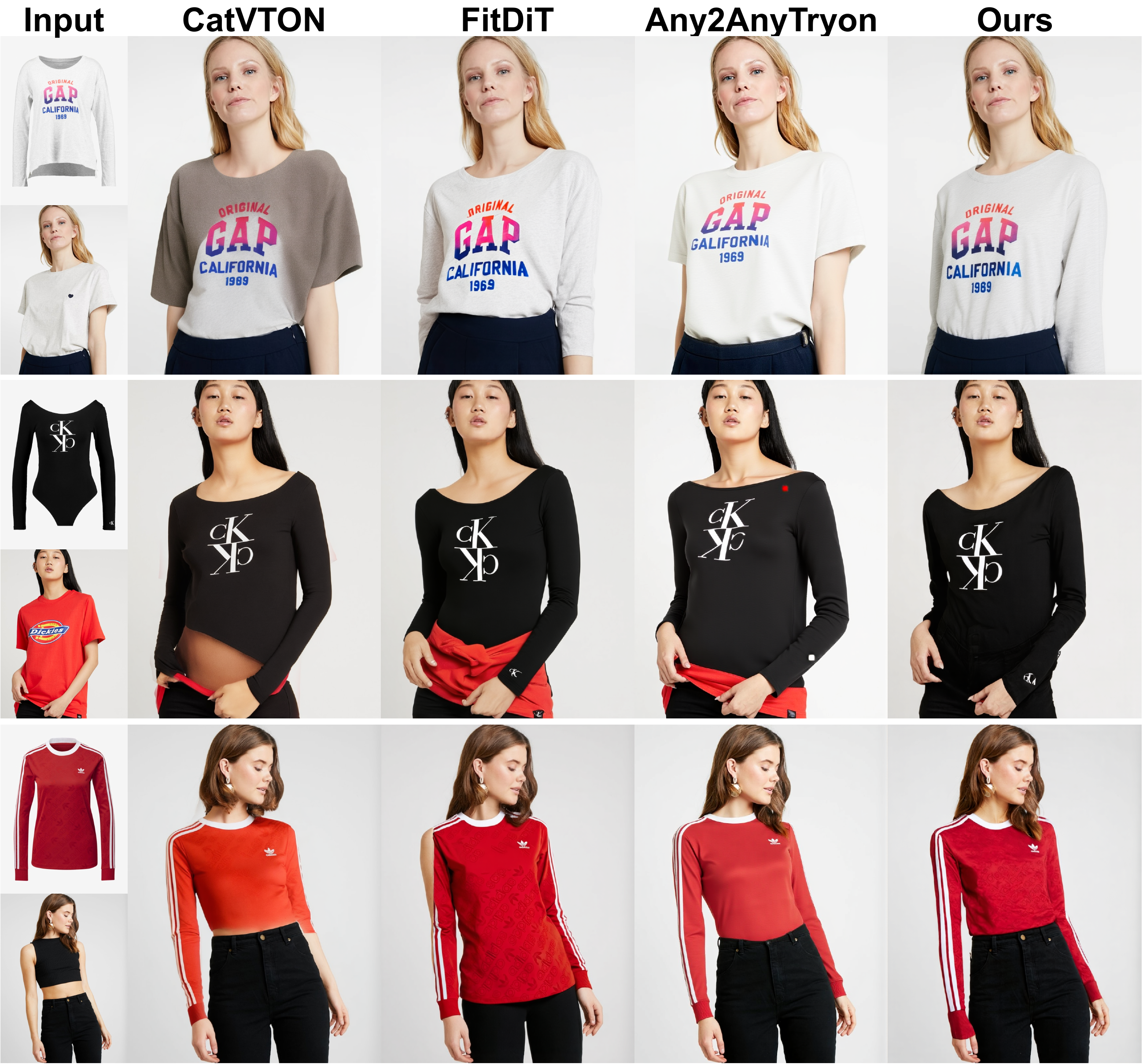}
\caption{Qualitative comparison of single-garment try-on.}
\label{fig tryon}
\end{figure}

\section{Experiment}
\subsection{Experimental Setup}
\subsubsection{Datasets} Our training incorporates a combination of public and self-synthesized data. For the three foundational tasks (single-garment try-on, model-free try-on, and garment reconstruction), we utilize the VITON-HD and DressCode datasets, which together provide approximately 59K training pairs for each task. For the advanced tasks, we leverage our self-synthesis pipeline: for multi-garment try-on, we generate 10K data pairs by reconstructing garments from DressCode; for model-to-model try-on, we synthesize 30K data pairs from both VITON-HD and DressCode. For the multi-view try-on task, we use a combination of the MVG and IG-Pairs datasets, totaling 12K samples. Textual instructions for all tasks are produced with Qwen2.5‑VL‑7B‑Instruct. Following existing mask‑free approaches~\cite{guo2025any2anytryon}, we employ Flux.1 Fill~\cite{flux} to synthesize triplet training samples. Pose images are extracted using DWPose~\cite{yang2023effective}. To obtain the ground-truth spatial mask for computing the spatial attention focusing loss, we employ SegFormer~\cite{xie2021segformer} to generate the corresponding garment segmentation.

\subsubsection{Training Details} The MGSA is built upon Qwen2‑VL‑2B‑Instruct. During training, we freeze its first 14 transformer layers and fine‑tune the remaining 14. We directly reuse the ViT embedded in Qwen2‑VL‑2B‑Instruct to extract target image features; this ViT processes \(756\times504\) inputs and outputs 486 visual tokens. Accordingly, we set the number of learnable queries to 486, each with a dimensionality of 1,536 to match the hidden size of Qwen2‑VL‑2B‑Instruct. For the DiT backbone, we experiment with StableDiffusion-3.5 Medium~\cite{esser2024scaling}. Our training process is divided into two stages. We first pre-train the model on the three foundational tasks for 120K steps. The training resolution is $1,024 \times 768 $ (or $768\times 576$ for model-free try-on), with a batch size of 16. We then conduct joint fine-tuning on all six tasks for another 80K steps. The resolutions are $1024 \times 768 $ for most tasks, and $768\times 576$ for model-free and multi-view try-on, with the same batch size of 16. Throughout all training stages, we use the AdamW optimizer with a learning rate of $4\times10^{-5}$ and set the gradient clipping threshold to 1.0.

\begin{figure}[t]
\centering
\includegraphics[width=0.95\columnwidth]{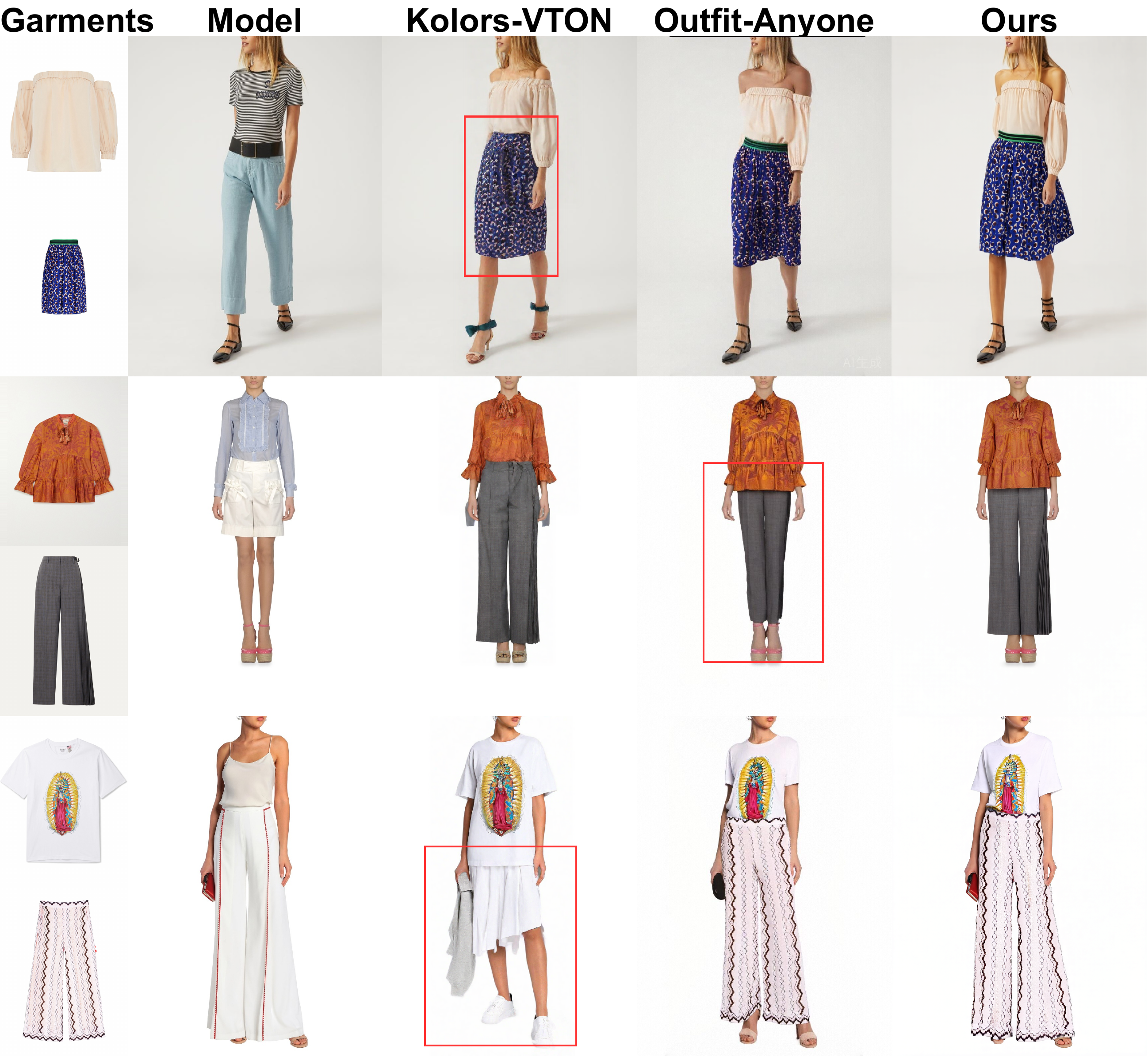}
\caption{Qualitative comparison of multi-garment try-on.}
\label{fig multi-garment}
\end{figure}

\begin{table}[t]
\begin{small}
\setlength{\tabcolsep}{5.5pt}
\centering
\begin{tabular}{lccc}
\toprule
\textbf{Method} & \textbf{CLIP-AS~$\uparrow$} & \textbf{CLIP-I~$\uparrow$} & \textbf{MP-LPIPS~$\downarrow$} \\
\midrule
IMAGDressing-v1  & \textbf{4.96} & 0.880 & 0.107 \\
Any2AnyTryon     & 4.95 & 0.843 & 0.127 \\
Ours   & 4.91 & \textbf{0.914} & \textbf{0.078} \\
\bottomrule
\end{tabular}
\caption{Quantitative comparison for model-free try-on on the VITON-HD dataset.}
\label{tab-model-free}
\end{small}
\end{table}

\subsection{Comparative Experiments}

\subsubsection{Garment Reconstruction}
We first evaluate our method on the garment reconstruction task, which involves reconstructing a flattened garment image from an image of a model wearing the target garment. To assess performance, we adopt four standard metrics: SSIM~\cite{wang2004image}, LPIPS~\cite{zhang2018unreasonable}, DISTS~\cite{ding2020image}, and
FID~\cite{heusel2017gans}. We compare the proposed UniFit with two state-of-the-art baselines—Try-OffDiff and Any2AnyTryon—on both the VITON-HD and DressCode datasets. The results, shown in Figure~\ref{fig Garment Reconstruction} and Table~\ref{Garment Reconstruction}, clearly demonstrate the superiority of our approach. Qualitatively, TryOffDiff often fails to preserve essential visual attributes such as color and texture, while Any2AnyTryon—an instruction-guided multi-task VTON framework—frequently struggles to comply with task instructions. For instance, as seen in the second row of Figure~\ref{fig Garment Reconstruction}, it fails to reconstruct the garment specified by the textual instruction. In contrast, UniFit accurately follows the textual instructions to reconstruct garments that closely match those worn by the model in the reference image, both in shape and fine-grained patterns. Quantitatively, UniFit achieves state-of-the-art performance on the VITON-HD dataset, significantly surpassing both baselines across multiple perceptual and distributional metrics. The strong improvements over Any2AnyTryon further highlight UniFit's ability to generate high-fidelity and semantically aligned results under textual instruction guidance.

\begin{figure}[t]
\centering
\includegraphics[width=0.95\columnwidth]{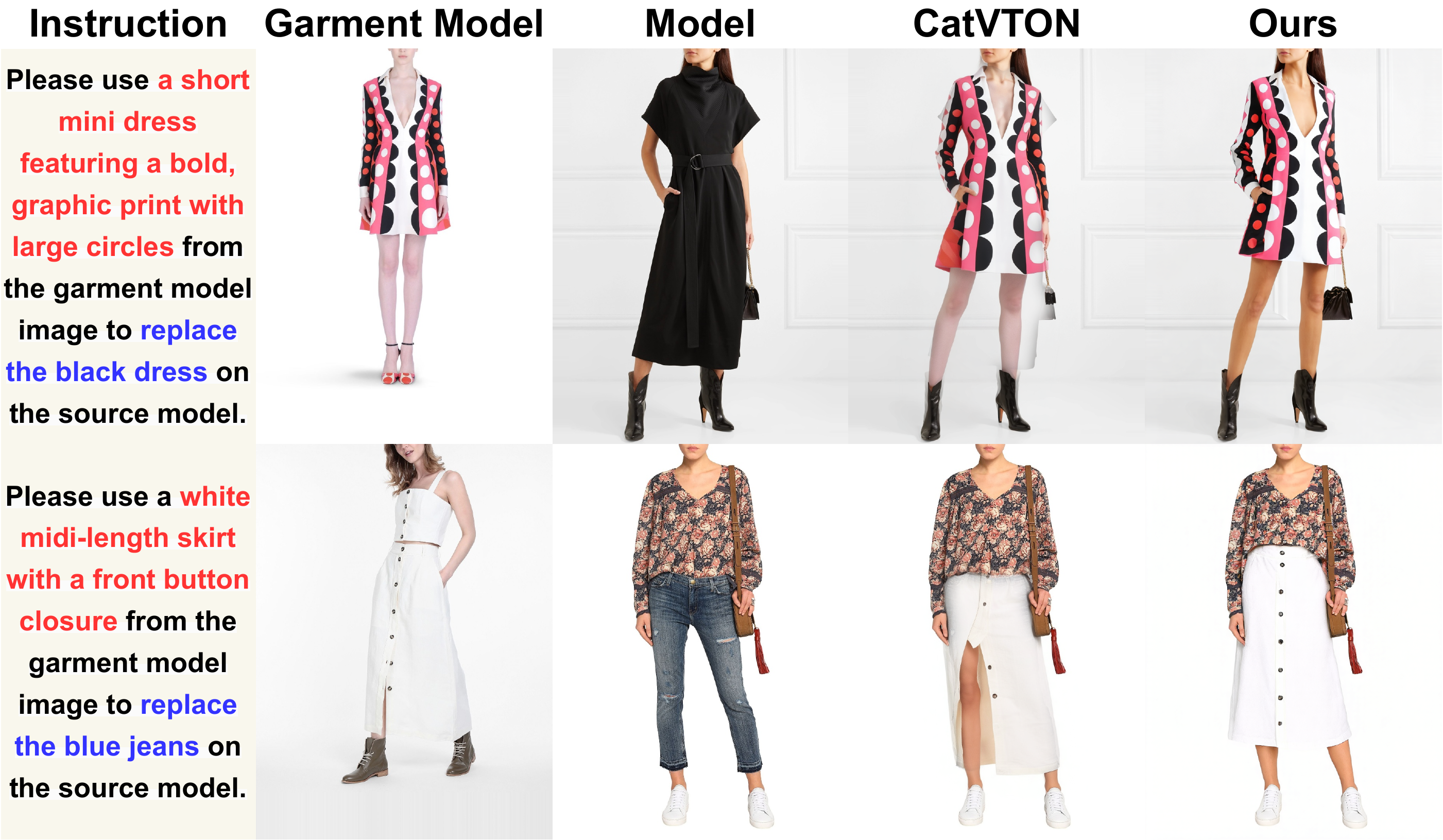}
\caption{Qualitative comparison of model-to-model try-on.}
\label{fig model2model}
\end{figure}

\subsubsection{Single-Garment Try-On}
We compare our method with CatVTON, FitDiT~\cite{jiang2024fitdit}, and Any2AnyTryon on the single-garment try-on task. As reported in Table~\ref{tryon}, UniFit outperforms all baselines across multiple metrics, with notable improvements in FID and KID~\cite{binkowski2018demystifying}, demonstrating superior generation quality. Figure~\ref{fig tryon} provides qualitative comparisons, where UniFit not only generates higher-quality and more realistic outfitted model images but also ensures that the garments worn by the model align closely with the input garments.

\begin{table}[t]
\begin{small}
\setlength{\tabcolsep}{9.5pt}
\centering
\begin{tabular}{lcccc}
\toprule
\textbf{Method} & \textbf{SSIM $\uparrow$} & \textbf{LPIPS $\downarrow$} & \textbf{FID $\downarrow$} & \textbf{KID $\downarrow$} \\
\midrule
MV-TON & 0.930 & \textbf{0.062} & 37.09 & \textbf{3.23} \\
Ours   & \textbf{0.935} & 0.072 & \textbf{35.62} & 3.85 \\
\bottomrule
\end{tabular}
\caption{Quantitative results on the MVG dataset for multi-view try-on.}
\label{tab-multiview}
\end{small}
\end{table}

\begin{figure}[t]
\centering
\includegraphics[width=0.95\columnwidth]{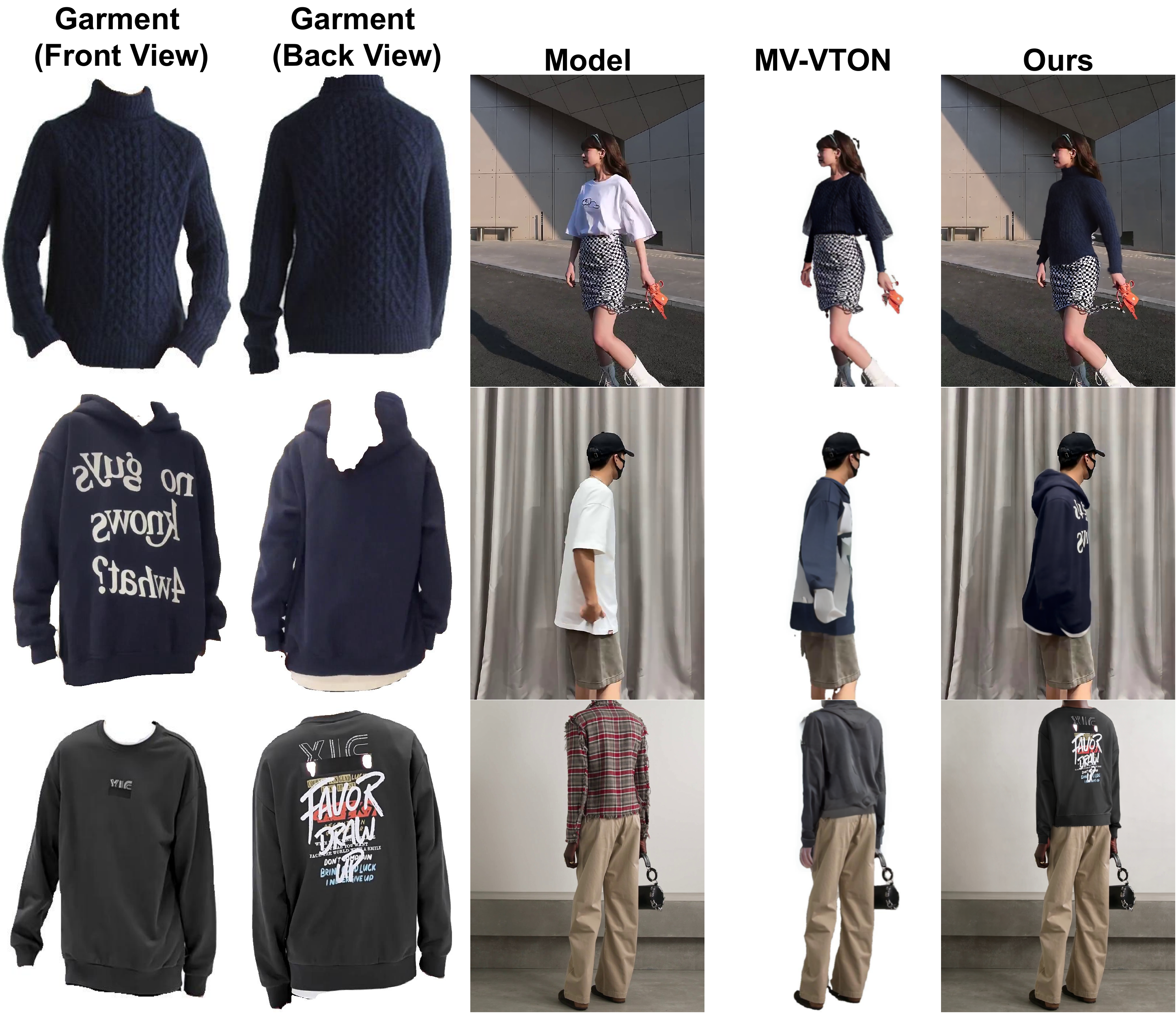}
\caption{Qualitative comparison of multi-view try-on.}
\label{fig-multi-view}
\end{figure}

\subsubsection{Model-Free Virtual Try-On}
Model-free virtual try-on refers to garment-driven model generation and can be viewed as a specific sub-task of subject-driven image generation. We evaluate UniFit on this task using the VITON-HD dataset and compare it against IMAGDressing-v1~\cite{shen2025imagdressing} and Any2AnyTryon. Three metrics are used for evaluation: CLIP-I~\cite{radford2021learning}, MP-LPIPS~\cite{chen2024magic}, and CLIP-AS~\cite{schuhmann2022laion}. Among them, MP-LPIPS is particularly effective at measuring the visual consistency between the input garment and the garment rendered on the generated model. As shown in Table~\ref{tab-model-free}, UniFit achieves superior performance in both MP-LPIPS and CLIP-I, while maintaining competitive results in CLIP-AS. These results indicate that our method not only produces high-quality model images but also better preserves the visual appearance of the input garments, demonstrating strong consistency and realism.

\subsubsection{Multi-garment Try-on}Most existing VTON methods are restricted to single-garment scenarios and lack the capability to flexibly compose and render combinations of tops and bottoms. Due to the absence of open-source multi-garment baselines, we conduct a qualitative comparison with two closed-source tools—OutfitAnyone~\cite{sun2024outfitanyone} and Kolors-VTON~\cite{team2024kolors}. As illustrated in Figure~\ref{fig multi-garment}, our method demonstrates superior ability in preserving the shape and details of garments. For instance, Kolors-VTON fails to maintain the skirt texture in the first row and introduces noticeable distortion in the third-row outfit. OutfitAnyone, on the other hand, often alters the garment structure, such as the pants in the second row. In contrast, UniFit produces more realistic and coherent try-on results, faithfully retaining the original garment appearance and silhouette across various combinations.

\subsubsection{Model-to-model Try-on}
Model-to-model Try-on aims to transfer specific garments from a source model to a target model. This task demands both accurate comprehension of the instruction and precise localization of the garments to be transferred. While existing methods such as CatVTON rely on explicit garment masks to guide the transfer, UniFit operates in a fully instruction-driven manner without requiring such auxiliary inputs. As shown in Figure~\ref{fig model2model}, qualitative results on the DressCode dataset indicate that UniFit can reliably interpret textual instructions, accurately identify the target garments, and generate high-fidelity try-on results with improved coherence and realism.

\subsubsection{Multi-view Try-on}
Multi-view Try-on aims to generate realistic try-on results by utilizing both the front and back views of the garment. MV-VTON is the first to explore this task, introducing a pose-aware hard-selection and soft-selection mechanism to guide the aggregation of garment features across views. In contrast, our method capitalizes on the cross-modal reasoning capability of the proposed MGSA module, which enables efficient semantic integration of multi-view garment features guided by instructions. As shown in Figure~\ref{fig-multi-view}, UniFit effectively fuses visual cues from different viewpoints to produce high-quality try-on results. Notably, even under challenging viewing angles (e.g., the side view in the second row), our method maintains garment realism and structural consistency. Furthermore, as demonstrated in Table~\ref{tab-multiview}, UniFit achieves superior performance in both fidelity and perceptual quality metrics, confirming the effectiveness of our approach in multi-view scenarios.

\begin{table}[t]
\begin{small}
\setlength{\tabcolsep}{8pt}
\centering
\begin{tabular}{lcccc}
\toprule
\textbf{Method} & \textbf{SSIM} $\uparrow$ & \textbf{LPIPS} $\downarrow$ & \textbf{FID} $\downarrow$ & \textbf{KID} $\downarrow$ \\
\midrule
w/o MGSA  & 0.851 & 0.098 & 9.133 & 1.053 \\
w/o $\mathcal{L}_{\text{align}}$ & 0.863 & 0.074 & 8.937 & 0.951 \\
w/o $\mathcal{L}_{\text{focus}}$ & 0.872 & \textbf{0.069} & 8.870 & 0.835 \\
Ours (Stage I) & \textbf{0.887} & 0.071 & \textbf{8.813} & \textbf{0.785} \\
\bottomrule
\end{tabular}

\caption{Quantitative results of ablation study on the VITON-HD dataset.}
\label{ab}
\end{small}
\end{table}
\begin{figure}[t]
\centering
\includegraphics[width=0.95\columnwidth]{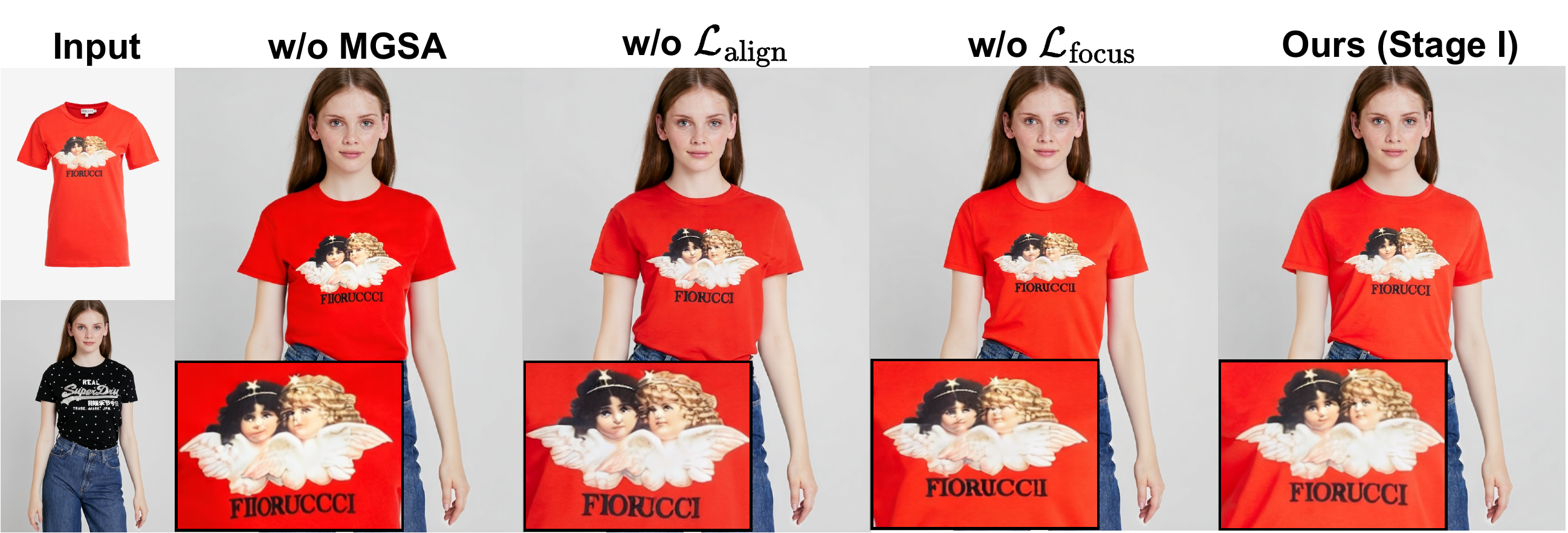}
\caption{Qualitative results of ablation study on the VITON-HD dataset.}
\label{fig-ab-tryon}
\end{figure}

\subsection{Ablation Study}
To demonstrate the effectiveness of our key components—the MGSA and the spatial attention focusing loss—we conduct an ablation study on the single-garment try-on task. To ensure fair comparison, all ablation studies are performed on our Base Model trained only through Stage I, thus avoiding any potential biases introduced by the self-synthesis process. As shown in Table~\ref{ab} and Figure~\ref{fig-ab-tryon}, we evaluate three variants: (1) w/o MGSA, where we replace the MGSA module (including $\mathcal{L}_{\text{align}}$) with a T5 text encoder; (2) w/o $\mathcal{L}_{\text{align}}$, where the semantic alignment loss is removed; and (3) w/o $\mathcal{L}_{\text{focus}}$, where the spatial attention focusing loss is removed. The results clearly show that both ablated models exhibit a noticeable degradation in performance compared to the full model, which confirms the effectiveness and necessity of our proposed components.

\section{Conclusion}
In this paper, we present UniFit, an instruction-guided universal VTON framework capable of flexibly handling a diverse range of complex tasks. Our work effectively addresses two core challenges faced by existing instruction-guided multi-task VTON frameworks: (1) the semantic gap between textual instructions and reference images, and (2) the data scarcity for complex scenarios. Our solution centers on two key innovations. First, an MLLM-Guided Semantic Alignment Module leverages a multimodal large language model to capture semantic relationships between instructions and reference images. It produces coherent and explicit semantic guidance for the generative process, effectively bridging the semantic gap. Second, a two-stage progressive training strategy with a self-synthesis pipeline enables UniFit to learn advanced tasks from limited data, overcoming the limitations of current datasets. Extensive experiments demonstrate that UniFit achieves state-of-the-art performance across six diverse VTON tasks, from single-garment try-on to complex multi-garment and model-to-model try-on scenarios.

\section{Limitations}
Despite achieving impressive performance across multiple VTON tasks, UniFit is constrained by the distribution of existing datasets, which predominantly feature in-shop scenarios. Consequently, there is a risk of performance degradation in in-the-wild settings, particularly under conditions of extreme lighting and severe occlusion. Furthermore, limited by the current mask-free strategy and the scarcity of task-specific training data, UniFit currently does not support virtual try-on in layers or text-editable virtual try-on.
\bibliography{aaai2026}

\clearpage
\twocolumn[
    \centering
    \Large \textbf{Appendix} % 标题
    \vspace{0.5em}           % 标题和正文的间距
    \vspace{1em}             % 再多留点空
] 
\appendix

This supplementary material serves as an extension to our main paper, offering additional details on our method, implementation, and experiments. We begin by reviewing related work. Following this, we introduce the preliminaries and the final optimization objective of our method, along with a visual illustration of the input and output formats for different tasks during inference. We then elaborate on the data processing pipeline, including the generation of textual instructions and the filtering strategy applied to our self-synthesized data. To further validate the effectiveness of each component in our framework, we provide additional results from ablation studies. Finally, we provide extended qualitative comparisons and report experiment results using Flux.1 Fill~\cite{flux} as UniFit’s DiT backbone.

\section{Related Work}
Image-based virtual try-on (VTON) aims to generate photorealistic images of a person wearing a specified garment while faithfully preserving the garment’s texture and the person’s identity. Early methods~\cite{xing2022virtual,xie2023gp,choi2021viton} typically followed a two-stage pipeline: first, a warping module aligned the garment to the target person’s pose; then, a GAN-based generator blended the warped garment onto the person’s image. However, this paradigm suffered from several limitations. Inaccuracies in the warping stage often introduced visual artifacts, and the instability of GAN training compromised both realism and texture fidelity. The advent of pre-trained diffusion models~\cite{rombach2022high,esser2024scaling,team2024kolors} has marked a paradigm shift in VTON, enabling substantial improvements in image quality and generation fidelity. Some methods, such as LaDI-VTON~\cite{morelli2023ladi}, refine the conventional two-stage pipeline by replacing the GAN-based generator with a diffusion-based generator, resulting in more realistic and high-fidelity outputs. Going further, methods like OOTDiffusion~\cite{xu2025ootdiffusion} and IDM-VTON~\cite{choi2024improving} eliminate explicit warping altogether. These approaches utilize parallel U-Net architectures and attention mechanisms to directly align features between garments and the target person, effectively reducing artifacts and improving alignment. To enhance spatial consistency and detail preservation, Leffa~\cite{zhou2024learning} introduces attention flow regularization. Meanwhile, lightweight models such as CatVTON~\cite{chong2024catvton} simplify the dual U-Net architecture using spatial concatenation, achieving competitive performance with fewer parameters. Despite these advances, most diffusion-based VTON methods~\cite{jiang2024fitdit,zhang2024boowvtonboostinginthewildvirtual,luan2025mcvtonminimalcontrolvirtual} remain task-specific and lack flexibility. Although recent efforts~\cite{zhang2024mmtryon,guo2025any2anytryon} have begun exploring instruction-guided multi-task VTON frameworks, they often face two fundamental challenges: (1) a semantic gap between abstract textual instructions and visual references, which leads to inconsistent or low-quality outputs; and (2) the data scarcity for complex scenarios, which hinders support for complex scenarios such as multi-garment try-on or model-to-model try-on.

\begin{figure*}[htp]
\centering
\includegraphics[width=0.8\textwidth]{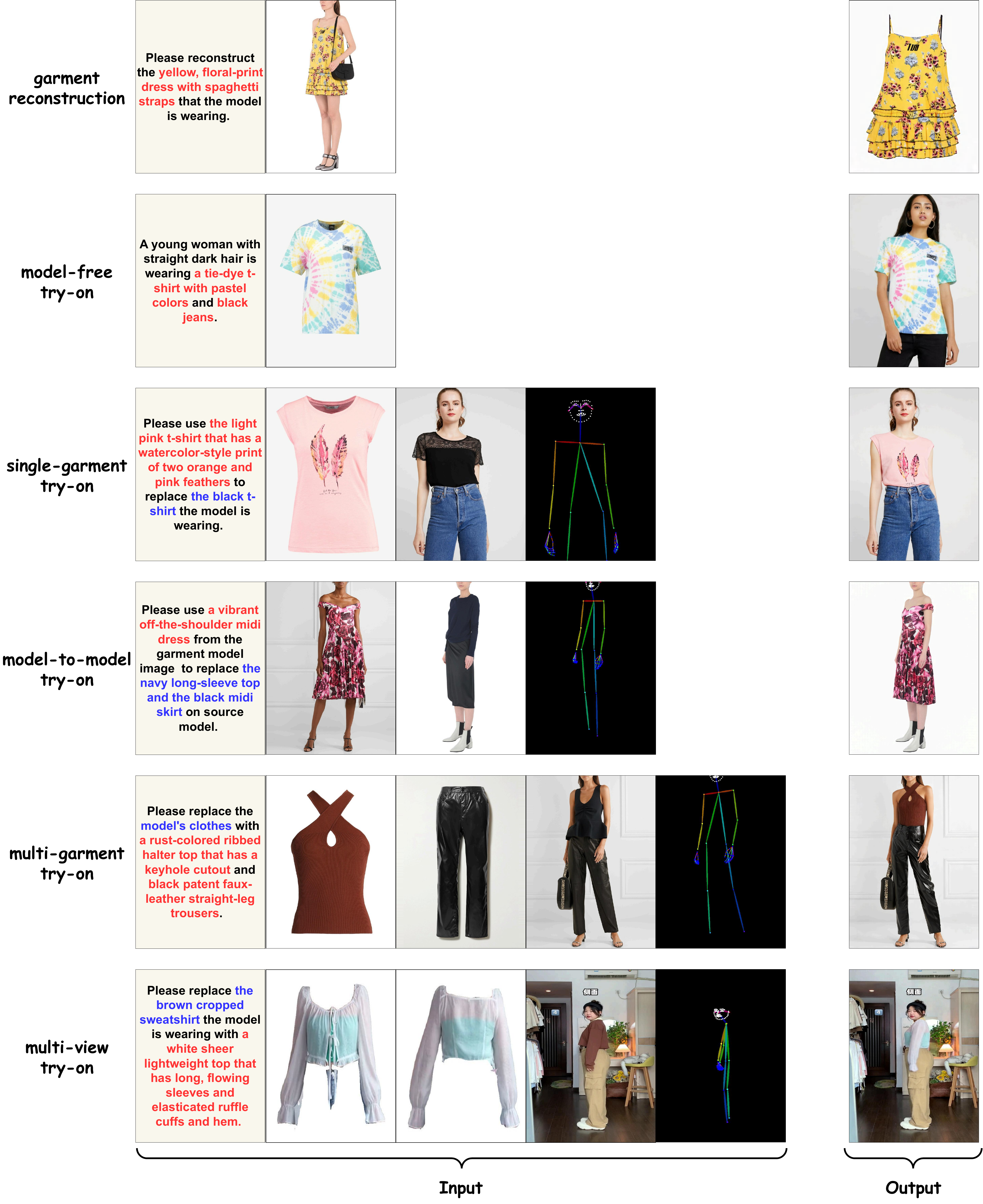}
\caption{Illustration of input and output formats across the six supported VTON tasks during inference. For each task, we show the textual instruction (middle-left), visual inputs (left), and the corresponding output image (right).}
\label{fig task input output}
\end{figure*}

\begin{figure*}[htp]
\centering
\includegraphics[width=0.8\textwidth]{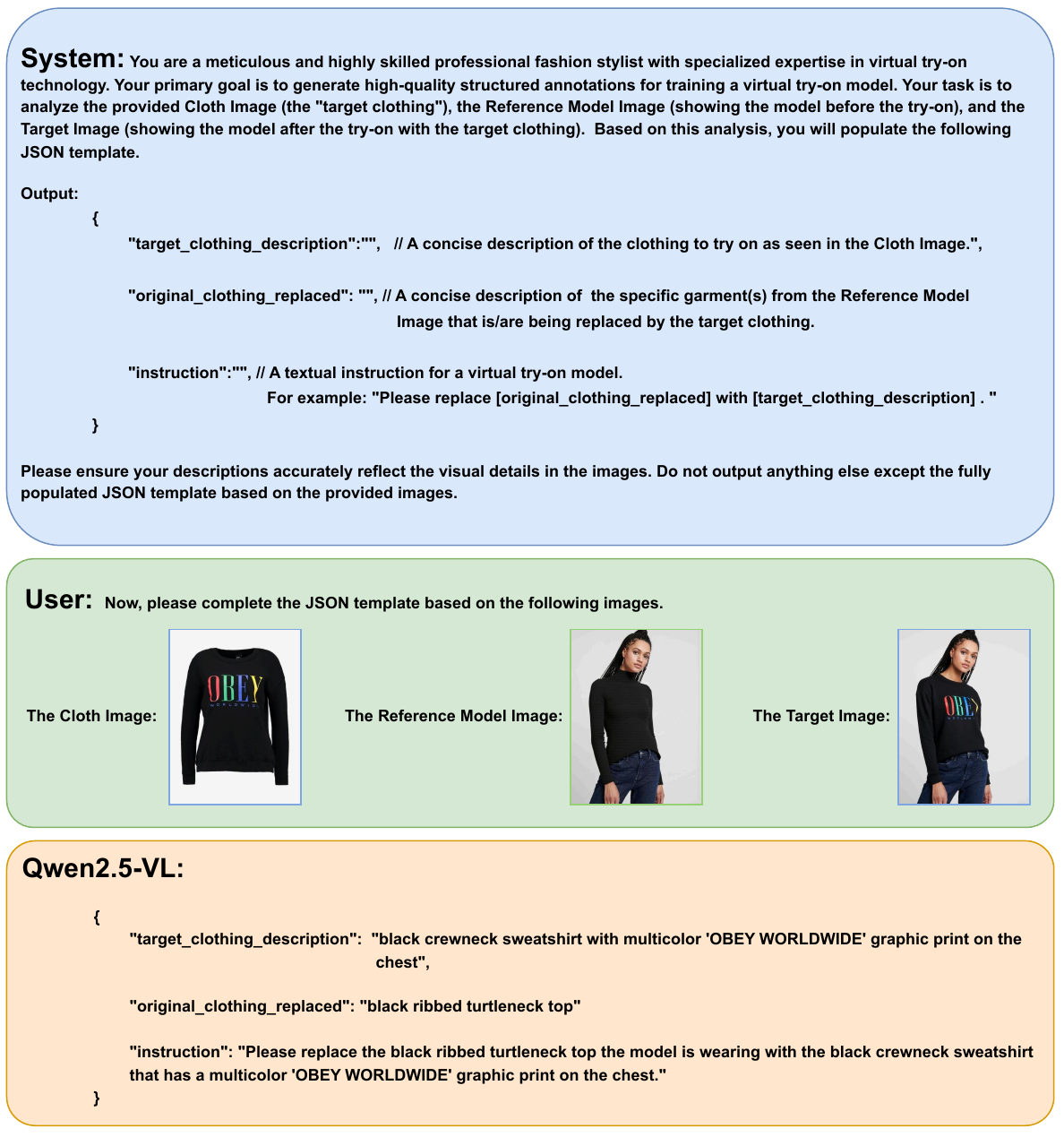}
\caption{Example of instruction generation for the single-garment try-on task using Qwen2.5‑VL‑7B‑Instruct.}
\label{fig instruction-gen}
\end{figure*}

\section{Method Details}

\subsection{Preliminaries}
Our generative model is built upon the Multi-modal Diffusion Transformer (MMDiT)~\cite{esser2024scaling}, which extends the Diffusion Transformer (DiT)~\cite{peebles2023scalable} framework by processing both image and text inputs as a unified sequence of tokens. Specifically, a noisy image latent $\mathbf{z}_t$ is first divided into a grid of patches, which are then flattened into a sequence of tokens. These image tokens are concatenated with other conditional embeddings and jointly processed by the MMDiT architecture. MMDiT is trained using Flow Matching~\cite{lipman2023flowmatchinggenerativemodeling}, an efficient alternative to traditional diffusion training. It defines the forward process via a linear interpolation between data and noise as $\mathbf{z}_t = t \mathbf{z}_0 + (1 - t) \boldsymbol{\epsilon}$, where $\mathbf{z}_0$ is the clean latent and $\boldsymbol{\epsilon} \sim \mathcal{N}(0, 1)$ is Gaussian noise. The model learns to predict the target velocity given the noised latent $\mathbf{z}_t$, timestep $t$, and conditional input $y$ (e.g., text or reference image). The training objective minimizes the mean squared error (MSE) between the ground-truth and predicted velocities:
\begin{equation}
    \mathcal{L}_{\text{diff}} = \mathbb{E} \left[\left\| (\mathbf{z}_0 - \boldsymbol{\epsilon}) - \mathcal{V}_\theta(\mathbf{z}_t, t, y) \right\|^2\right],
\end{equation}
where $\mathcal{V}_\theta$ is the diffusion model.

\begin{table*}[h]
    \centering
    \small
    \begin{tabular}{lcccccc} 
        \toprule
        \multirow{2}{*}{\textbf{Method}} & \multirow{2}{*}{\textbf{Backbone}} & \multirow{2}{*}{\textbf{Total Params}} & \multicolumn{2}{c}{\textbf{Single-Garment}} & \multicolumn{2}{c}{\textbf{Multi-Garment (2 items)}} \\
        \cmidrule(lr){4-5} \cmidrule(lr){6-7}
        & & & \textbf{VRAM} & \textbf{Time} & \textbf{VRAM} & \textbf{Time} \\
        \midrule
        FitDiT & SD3 & 5.6B & \textbf{16GB} & \textbf{6s} & - & - \\
        Any2AnyTryon & Flux & 17B & 36GB & 56s & - & - \\
        \textbf{UniFit (Ours)} & SD3.5M & \textbf{4.8B} & \textbf{16GB} & 14s & \textbf{18GB} & \textbf{22s} \\
        \bottomrule
    \end{tabular}
    \caption{Comparison of computational resources at $1024 \times 768$ resolution. UniFit achieves a balance between performance and efficiency, drastically outperforming the SOTA large-model baseline (Any2AnyTryon) in both speed and memory usage.}
    \label{tab-efficiency}
\end{table*}

\subsection{Optimization Objective}
As outlined in the main paper, the overall training objective integrates three loss components: (1) the flow matching loss $\mathcal{L}_{\text{diff}}$, (2) the semantic alignment loss $\mathcal{L}_{\text{align}}$, and (3) the spatial attention focusing loss $\mathcal{L}_{\text{focus}}$. The final objective function is formulated as:
\begin{equation}
    \mathcal{L} = \lambda_{\text{diff}} \cdot \mathcal{L}_{\text{diff}} + \lambda_{\text{align}} \cdot \mathcal{L}_{\text{align}} + \lambda_{\text{focus}} \cdot \mathcal{L}_{\text{focus}},
\end{equation}
where $\lambda_{\text{diff}}$, $\lambda_{\text{align}}$, and $\lambda_{\text{focus}}$ are weighting coefficients for each loss term. In our implementation, we set $\lambda_{\text{diff}} = 1$, $\lambda_{\text{align}} = 1$, and $\lambda_{\text{focus}} = 0.05$.

\subsection{Task Inputs During Inference}
To better illustrate the task-specific inputs during inference, Figure~\ref{fig task input output} presents the input and output formats for all six supported virtual try-on tasks in UniFit. Each row corresponds to a different task, where we visualize the task-specific textual instruction, associated visual inputs (such as garment images, model images, or pose images), and the final synthesized output image.

\section{Data Processing}
\subsection{Instruction Generation}
As described in the main paper, we leverage task-specific textual instructions to guide the model in performing various virtual try-on (VTON) tasks. To obtain these instructions, we utilize Qwen2.5‑VL‑7B‑Instruct~\cite{bai2025qwen2} to analyze the visual relationships between the input and output images of each task, thereby generating task-relevant textual annotations. Specifically, we first reorganize and categorize the training data based on task types (e.g., single-garment try-on). For each task, the corresponding image data along with predefined description templates are fed into Qwen2.5‑VL‑7B‑Instruct to generate textual instructions tailored to the specific task. An example of this instruction generation process for the single-garment try-on task is shown in Figure~\ref{fig instruction-gen}.

\subsection{Filtering Strategy for Synthesized Data}

To ensure the quality of our self-synthesized data, we employ a two-stage filtering strategy that combines low-level perceptual similarity assessment with high-level semantic verification.

\paragraph{Stage 1: Perceptual Similarity Filtering.} 
For multi-garment data pairs, we construct each pair by using the Base Model to reconstruct a specific garment image \( g_{\text{syn}} \) (e.g., a top or bottom) from a full-body reference model image \( x_{\text{ref}} \). Therefore, it is crucial to ensure that the synthesized garment \( g_{\text{syn}} \) faithfully represents the corresponding garment in the reference image. To assess reconstruction fidelity, we first apply a semantic segmentation model to extract the garment region \( g_{\text{seg}} \) from \( x_{\text{ref}} \) that corresponds to \( g_{\text{syn}} \). We then compute the perceptual similarity between \( g_{\text{seg}} \) and \( g_{\text{syn}} \) using DreamSim~\cite{fu2023dreamsim}. Pairs with similarity scores below a predefined threshold \( \tau \) are discarded. Similarly, for model-to-model data pairs, where a synthesized model image \( x_{\text{syn}} \) is generated based on a reference garment image \( g_{\text{ref}} \), we extract the garment region \( g_{\text{seg}} \) from \( x_{\text{syn}} \) and compute its similarity to \( g_{\text{ref}} \) using the same metric. Low-scoring pairs are removed in this stage as well. In practice, we set \( \tau = 0.35 \), and adopt SegFormer~\cite{xie2021segformer} as the garment segmentation model.

\paragraph{Stage 2: Semantic Consistency Filtering.}
In the second stage, we leverage Qwen2.5‑VL‑7B‑Instruct to assess the high-level semantic consistency between synthesized images and their corresponding reference images. For multi-garment data pairs, we input both the synthesized garment image and the reference model image into Qwen2.5‑VL‑7B‑Instruct, prompting the model to determine whether the garments in both images match. For model-to-model pairs, we similarly verify whether the synthesized model is wearing garments that are consistent with the reference garment image. Only data pairs that pass both stages are retained for training.

\section{Computational Efficiency Analysis}
\paragraph{Parameter Count.}
Our UniFit framework maintains a parameter count comparable to or significantly lower than existing state-of-the-art methods. Specifically, UniFit comprises the Qwen2-VL-2B (2.3B) and the Stable Diffusion 3.5 Medium backbone (2.5B), totaling approximately \textbf{4.8B} parameters. As shown in Table~\ref{tab-efficiency}, this is consistent with FitDiT (5.6B) and substantially more efficient than Any2AnyTryon, which utilizes a massive 17B parameter architecture.

\paragraph{Inference Costs (VRAM and Time).}
We measured the inference performance at a resolution of $1024 \times 768$ on a single NVIDIA A800 GPU. Table~\ref{tab-efficiency} presents the detailed comparison. In single-garment try-on tasks, while FitDiT achieves faster inference speeds by pruning cross-attention layers, this architectural simplification severely restricts its supported task range. In contrast, compared to Any2AnyTryon—which supports a similarly wide range of tasks—UniFit demonstrates a significant advantage in both VRAM efficiency and inference speed.

For multi-garment try-on, a critical challenge lies in the attention mechanism, where computational costs typically exhibit quadratic growth as more reference images are added. To address this, we incorporate the \textbf{Causal Conditional Attention} mechanism from EasyControl~\cite{zhang2025easycontrol} and UniCombine~\cite{wang2025unicombine} into our DiT backbone. This mechanism allows the model to process multiple reference conditions efficiently without a significant increase in memory consumption, ensuring that the resource overhead remains manageable even when handling multiple garments simultaneously.

\begin{table}[h]
    \centering
    \small
    \caption{User study results on the Street Tryon dataset. We report the percentage (\%) of votes where experts selected the method as the best. UniFit consistently achieves the highest preference rates across all metrics.}
    \label{tab:user_study}
    \setlength{\tabcolsep}{10pt} % 稍微调整列宽
    \begin{tabular}{lccc}
        \toprule
        Method & PBC (\%) & GC (\%) & OQ (\%) \\
        \midrule
        CatVTON & 9 & 8 & 8 \\
        FitDiT & 2 & 21 & 6 \\
        Kolors-VTON & 4 & 21 & 9 \\
        Any2AnyTryon & 17 & 22 & 15 \\
        \textbf{UniFit (Ours)} & \textbf{68} & \textbf{28} & \textbf{62} \\
        \bottomrule
    \end{tabular}
\end{table}

\section{User Study on In-the-Wild Scenarios}
To evaluate the robustness of UniFit in real-world, uncontrolled environments, we conducted a user study using the \textbf{Street Tryon} dataset. This dataset contains diverse images captured in complex settings with varying lighting and backgrounds, posing significant challenges for generalization.

\paragraph{Study Setup.}
We invited 10 professional fashion experts to participate in a blind evaluation, consisting of \textbf{9 test cases per participant}. For each test case, experts were presented with a source model image, a garment image, and the try-on results generated by UniFit and competitive baselines (e.g., CatVTON, FitDiT, Kolors-VTON~\cite{team2024kolors}, and Any2AnyTryon). The method names were anonymized (labeled A/B/C/D/E). Experts were asked to select the \textbf{best} result based on three specific criteria:
\begin{enumerate}
    \item \textbf{Person \& Background Consistency (PBC):} Measures the preservation of the source model's identity and the consistency of background elements.
    \item \textbf{Garment Consistency (GC):} Measures how well the generated garment matches the reference in terms of style, structure, and texture patterns.
    \item \textbf{Overall Quality (OQ):} A comprehensive assessment of aesthetics, realism, and natural integration (considering lighting, shadows, and edge transitions).
\end{enumerate}

\paragraph{Results.}
Table~\ref{tab:user_study} reports the percentage of votes where each method was selected as the best. UniFit demonstrates superior performance across all three dimensions. Notably, in terms of Overall Quality (OQ) and Person \& Background Consistency (PBC), UniFit secured \textbf{62\%} and \textbf{68\%} of the experts' votes respectively, highlighting its robustness in generating high-fidelity results in challenging in-the-wild scenarios. While the competition for Garment Consistency (GC) was closer, UniFit still achieved the highest preference rate (\textbf{28\%}).

\begin{figure}[t]
\centering
\includegraphics[width=0.85\columnwidth]{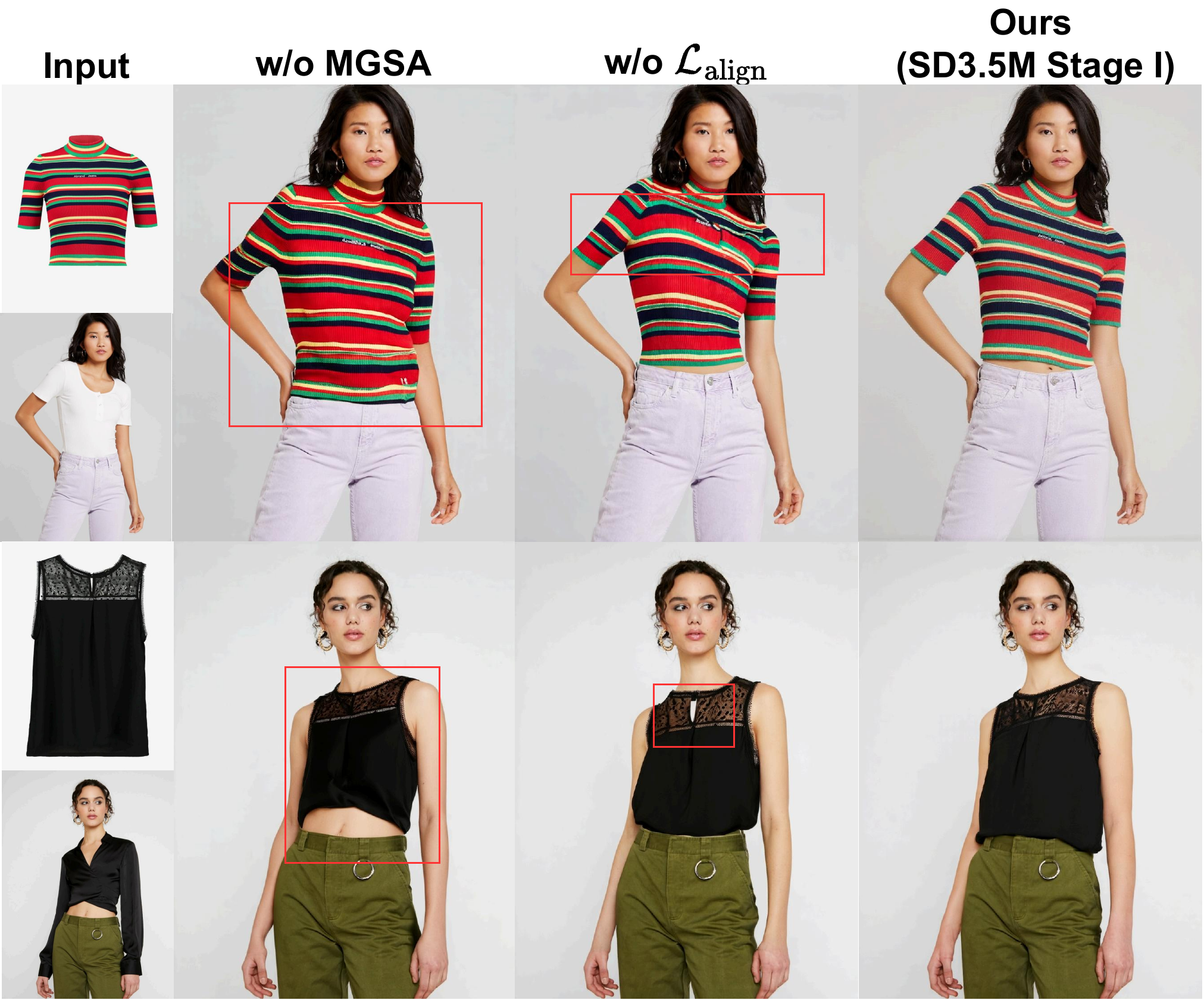}
\caption{ The ablation study of the MLLM-Guided Semantic Alignment Module (MGSA).}
\label{fig add ablation mgsa}
\end{figure}

\begin{figure}[t]
\centering
\includegraphics[width=0.85\columnwidth]{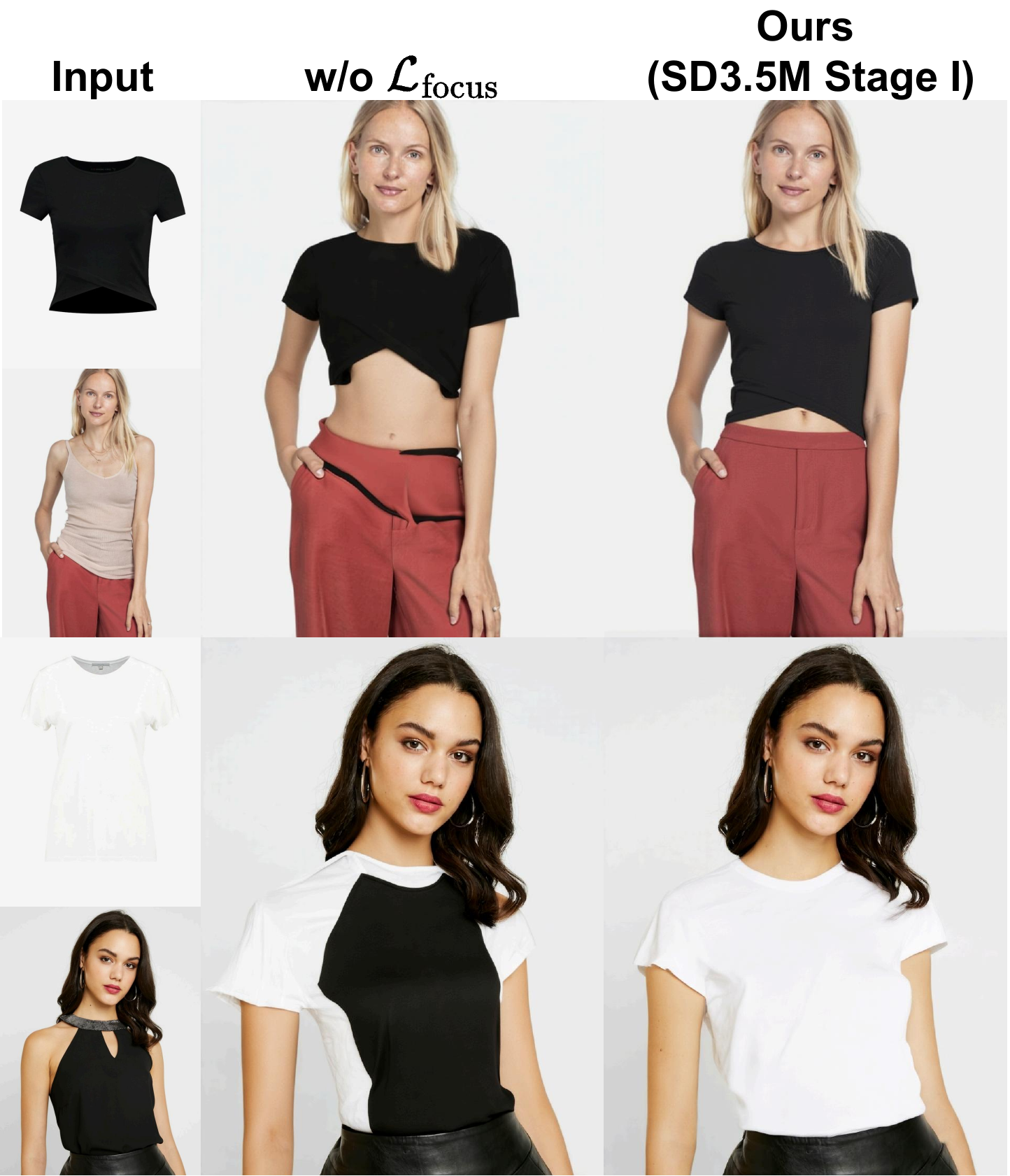}
\caption{ The ablation study of the spatial attention focusing loss. }
\label{fig add ablation focus}
\end{figure}

\section{Additional Ablation Results}

To further validate the effectiveness of key components in our framework, we conduct detailed ablation studies.

\paragraph{Effect of MGSA.} 
As illustrated in Figure~\ref{fig add ablation mgsa}, removing the MGSA module leads to significant shape inconsistencies between the generated garments and the reference garments, resulting in severe visual distortions. In addition, removing the semantic alignment loss $\mathcal{L}_{\text{align}}$ causes noticeable texture distortions and may introduce undesired artifacts, as shown in the second row of Figure~\ref{fig add ablation mgsa}. In contrast, the full UniFit model produces results with higher visual fidelity.

\paragraph{Effect of Focus Loss.}
Figure~\ref{fig add ablation focus} illustrates the effect of removing the spatial attention focus loss $\mathcal{L}_{\text{focus}}$. Without this loss, the model may produce artifacts such as color blending or boundary inconsistencies in the garment regions. By incorporating $\mathcal{L}_{\text{focus}}$, our model is guided to accurately transfer relevant features from different reference images to their corresponding target regions. This effectively reduces cross-image interference and results in more coherent and visually plausible try-on outcomes.

\section{Additional Comparative Experiments}
To further validate the effectiveness of our proposed method, this section presents additional qualitative comparisons. Specifically, in addition to showcasing results using Stable Diffusion 3.5 Medium~\cite{esser2024scaling} as the DiT backbone of UniFit, we also provide results based on Flux.1 Fill~\cite{flux} to demonstrate the scalability of our framework. For clarity, we refer to the two versions as SD3.5M and Flux Fill throughout this section.

\subsection{Training Setup}
The Flux Fill variant of UniFit adopts a training pipeline and architecture largely consistent with the SD3.5M version, with several key adjustments in training configurations. First, to reduce GPU memory consumption, we apply LoRA to fine-tune the attention layers of the DiT backbone, with the LoRA rank set to 128. Second, to accelerate training, we adjust both the training resolution and batch size. Specifically, in Stage I, the model is trained for 60K steps at a resolution of $512\times384$, followed by an additional 20K steps at a higher resolution—$1024\times768$ for most tasks, or $768\times576$ for model-free try-on. In Stage II, training continues at $1024\times768$ resolution for another 20K steps, except for model-free and multi-view try-on tasks, which are trained at a resolution of $768\times576$. Notably, in this stage, the Flux Fill variant is also trained using the self-synthesized data generated by the SD3.5M version. Throughout the entire training process, a batch size of 8 is used. All other hyperparameters are kept consistent with those of the SD3.5M variant. Both SD3.5M and Flux Fill versions are trained on 4 NVIDIA A800 (80GB) GPUs with DeepSpeed ZeRO-2 enabled for memory optimization.

\subsection{Additional Qualitative Results}

\begin{figure*}[ht]
\centering
\includegraphics[width=0.99\textwidth]{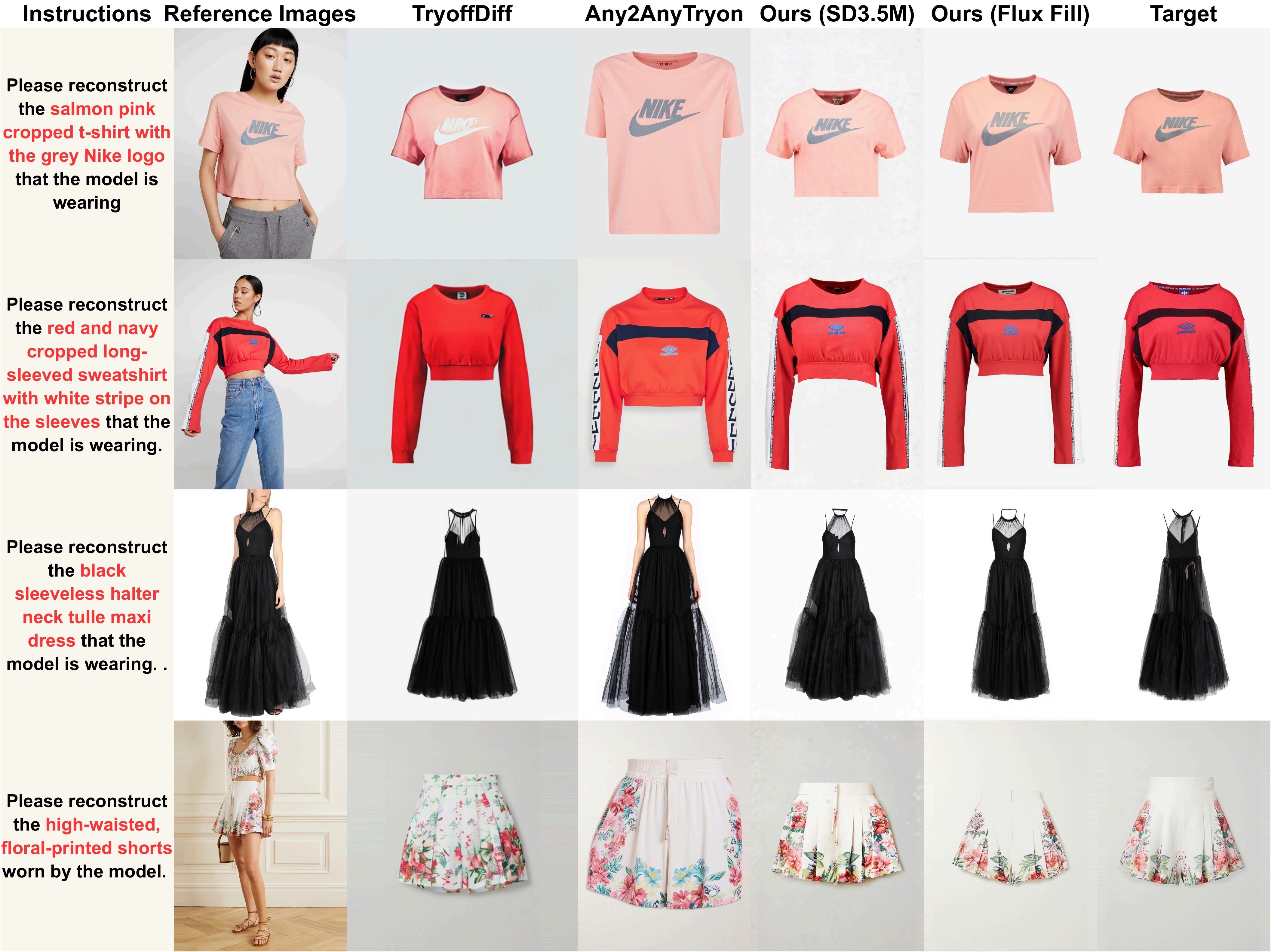}
\caption{More visual comparisons of garment reconstruction.}
\label{fig tryff}
\end{figure*}

\begin{figure*}[t]
\centering
\includegraphics[width=0.99\textwidth]{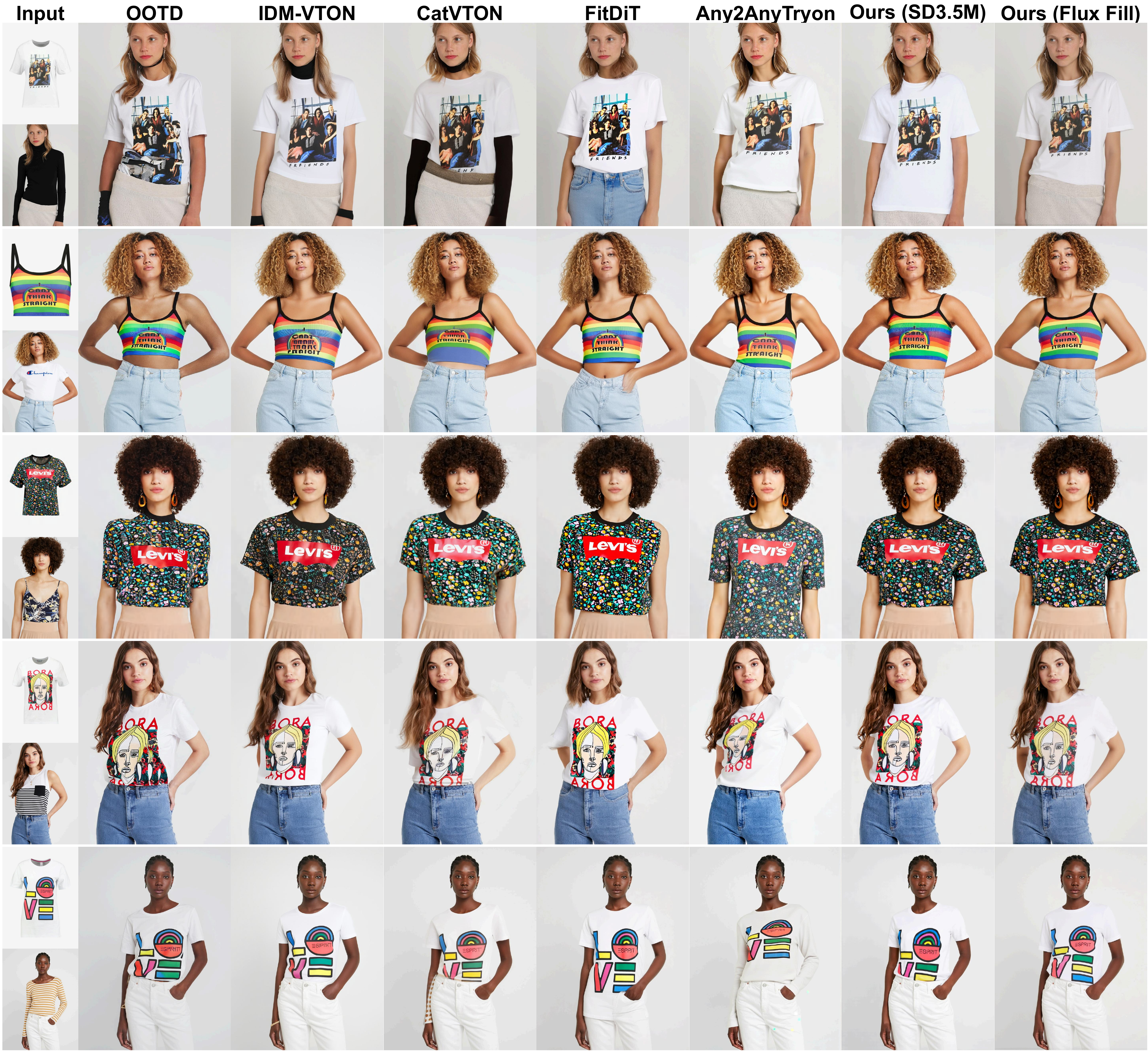}
\caption{More visual comparisons of single-garment try-on. Please zoom in for more details.}
\label{fig tryon appendix}
\end{figure*}

\begin{figure*}[t]
\centering
\includegraphics[width=0.99\textwidth]{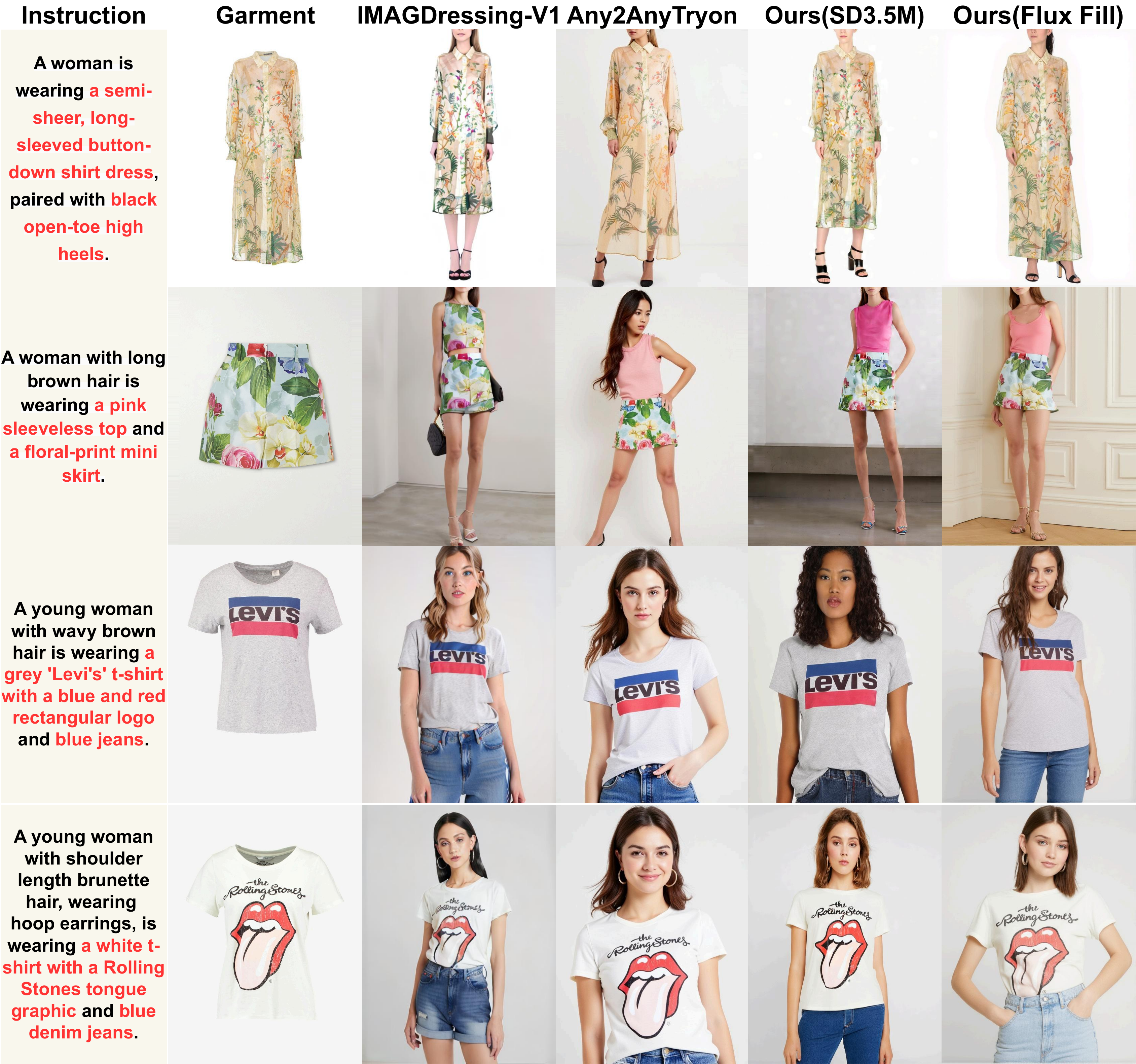}
\caption{More visual comparisons of model-free try-on.}
\label{fig model-free try-on}
\end{figure*}

\begin{figure*}[t]
\centering
\includegraphics[width=0.99\textwidth]{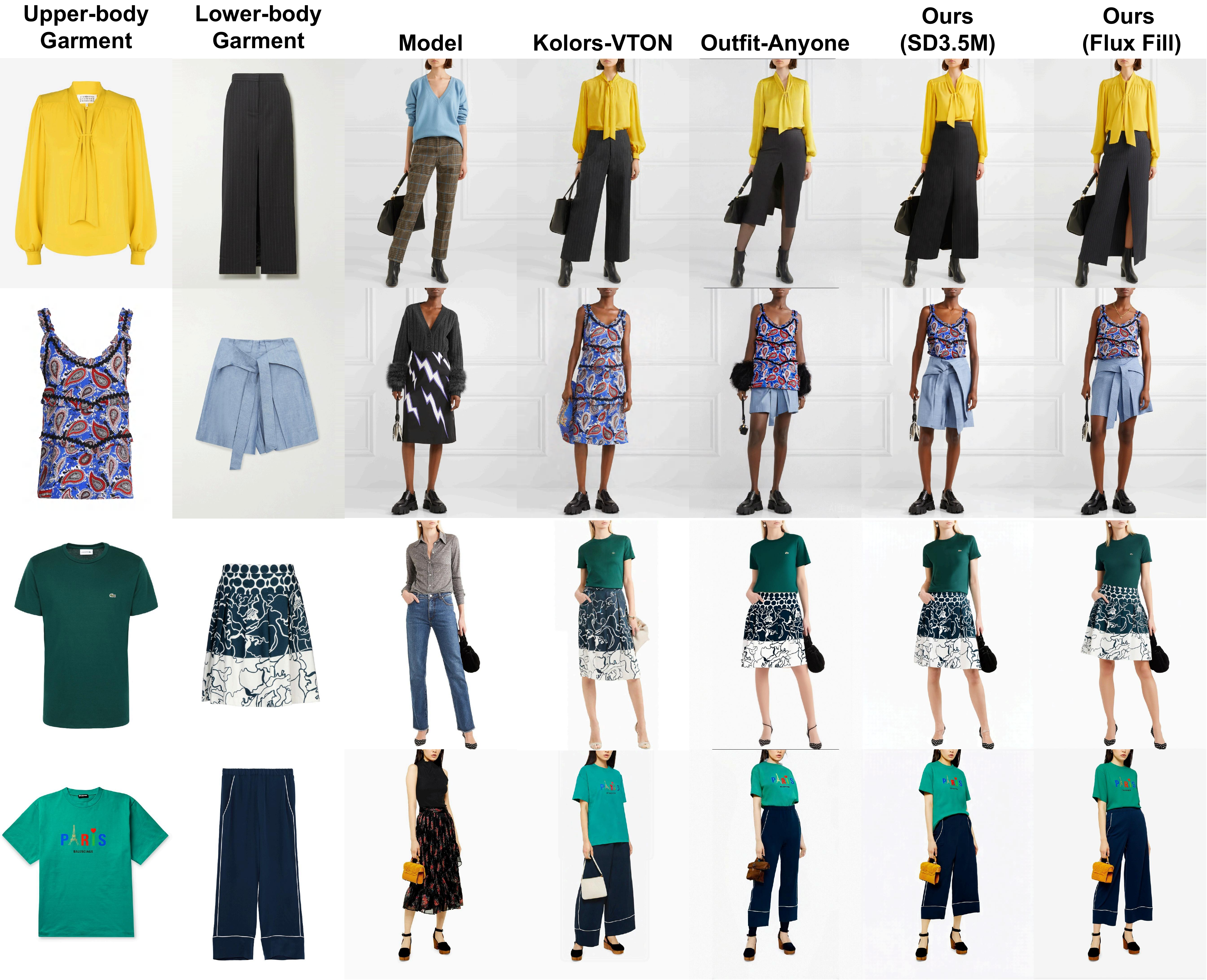}
\caption{More visual comparisons of  multi-garment try-on.}
\label{fig multi-garment try-on}
\end{figure*}

\begin{figure*}[t]
\centering
\includegraphics[width=0.99\textwidth]{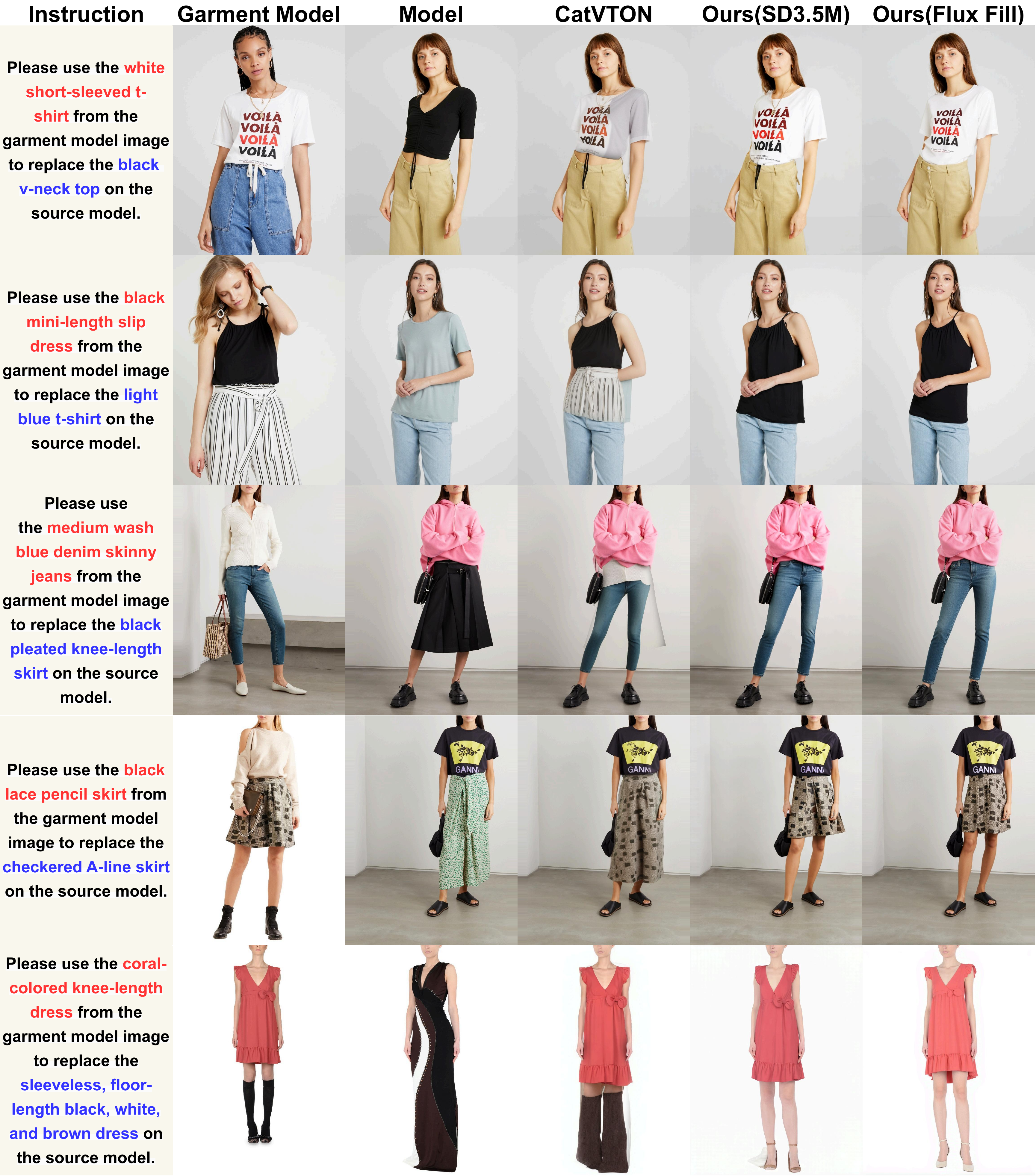}
\caption{More visual comparisons of  model-to-model try-on.}
\label{fig model-to-model try-on}
\end{figure*}

\begin{figure*}[t]
\centering
\includegraphics[width=0.99\textwidth]{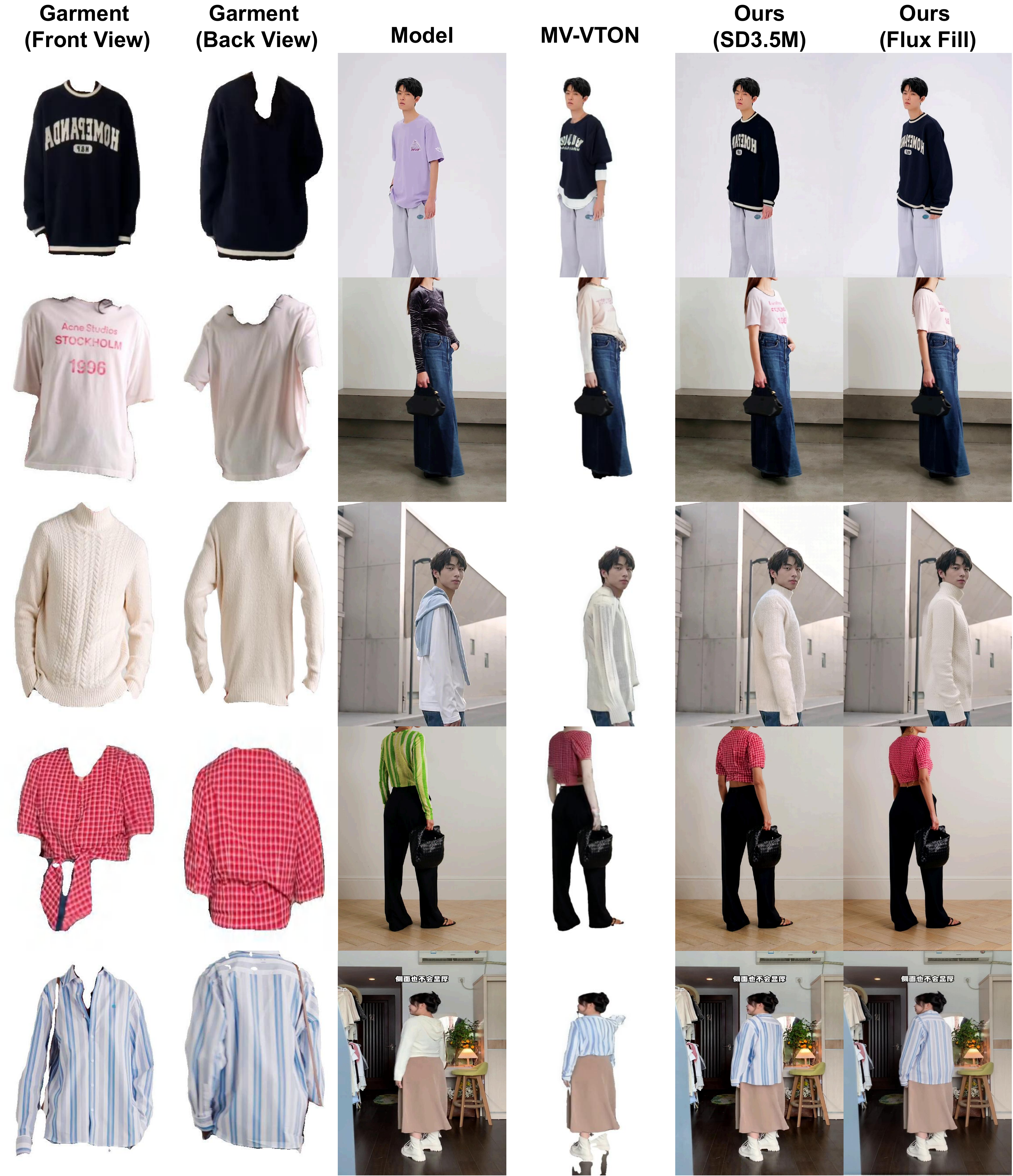}
\caption{More visual comparisons of multi-view try-on.}
\label{fig multi-view try-on}
\end{figure*}

Figures~\ref{fig tryff}, \ref{fig tryon appendix}, \ref{fig model-free try-on}, \ref{fig multi-garment try-on}, \ref{fig model-to-model try-on}, and \ref{fig multi-view try-on} present qualitative comparisons between the two UniFit versions (SD3.5M and Flux Fill) across the six supported VTON tasks. These results clearly demonstrate that our method can flexibly handle a wide range of try-on scenarios while generating visually realistic and high-quality outputs. While the SD3.5M version delivers comparable overall image quality, subtle differences emerge in certain fine-grained details. For example, as shown in the second row of Figure~\ref{fig multi-view try-on}, the Flux Fill version produces a more coherent rendering of the lower garment. We attribute this to the Flux Fill model’s larger parameter capacity and richer prior knowledge.

\end{document}